\documentclass[11pt,]{article}

\PassOptionsToPackage{hyphens}{url}

\usepackage[left=1in,top=1in,right=1in,bottom=1in]{geometry}
\usepackage{mathpazo}
\usepackage[T1]{fontenc}
\usepackage[utf8]{inputenc}
\usepackage{abstract}
\usepackage{amsmath}
\usepackage{longtable}
\usepackage{booktabs}
\usepackage{makecell}
\usepackage{array}
\usepackage{xcolor}
\usepackage{colortbl}
\usepackage{multirow}
\usepackage{graphicx}
\usepackage{caption}
\usepackage{rotating}
\usepackage{float}
\usepackage[section]{placeins}
\usepackage{csquotes}
\usepackage{setspace}
\usepackage{titlesec}
\usepackage[strings]{underscore}
\usepackage{natbib}
\usepackage{hyperref}
\usepackage{orcidlink}
\usepackage{listings}
\usepackage{sourcecodepro}
\usepackage{subcaption}

\newcommand*{\authorfont}{\fontfamily{ppl}\selectfont}

\newcolumntype{C}[1]{>{\centering\arraybackslash}p{#1}}

\definecolor{light-gray}{gray}{0.95}
\definecolor{steelblue}{rgb}{0.45,0.51,0.67}

\definecolor{codebg}{gray}{0.96}
\definecolor{codeframe}{gray}{0.80}
\definecolor{codecomment}{gray}{0.40}

\renewenvironment{abstract}
 {{%
    \setlength{\leftmargin}{0mm}
    \setlength{\rightmargin}{\leftmargin}%
  }%
  \relax}
 {\endlist}

\titleformat*{\section}{\normalsize\bfseries}
\titleformat*{\subsection}{\normalsize\bfseries\itshape}
\titleformat*{\subsubsection}{\normalsize\itshape}
\titleformat*{\paragraph}{\normalsize\itshape}
\titleformat*{\subparagraph}{\normalsize\itshape}

\hypersetup{
  breaklinks = true,
  bookmarks  = true,
  colorlinks = true,
  citecolor  = steelblue,
  filecolor  = steelblue,
  linkcolor  = steelblue,
  urlcolor   = steelblue,
  pdfborder  = {0 0 0},
  pdfauthor  = {Alexander Häußer},
  pdftitle   = {
    Echo State Networks for Time Series Forecasting:
    Hyperparameter Sweep and Benchmarking
  },
  pdfkeywords = {
    Echo State Networks,
    Reservoir Computing,
    Time Series Forecasting,
    Hyperparameter,
    Model Configuration,
    Forecasting Benchmarks,
    M4 Forecasting Competition
  }
}

\makeatletter

\def\@maketitle{%
  \newpage
  \let\footnote\thanks
  {%
    \fontsize{24}{26}\selectfont
    \raggedright
    \setlength{\parindent}{0pt}
    \bfseries
    \@title
    \par
  }%
}

\makeatother

\newcommand\extrafootertext[1]{%
  \bgroup
  \renewcommand\thefootnote{\fnsymbol{footnote}}%
  \renewcommand\thempfootnote{\fnsymbol{mpfootnote}}%
  \footnotetext[0]{#1}%
  \egroup
}

\makeatletter
\def\fps@figure{htbp}
\makeatother

\title{Echo State Networks for Time Series Forecasting: Hyperparameter
Sweep and Benchmarking}

\newcommand{\AuthorName}{Alexander Häußer}

\newcommand{\AuthorInstitution}{Justus-Liebig-University Giessen, Faculty
of Economics and Business Studies, Chair of Statistics and Econometrics}

\newcommand{\AuthorAddress}{Licher Strasse 64, 35394 Giessen, Germany}

\newcommand{\AuthorEmail}{alexander.haeusser@wirtschaft.uni-giessen.de}

\newcommand{\AuthorORCID}{0009-0000-5419-8479}

\newcommand{\AuthorFootnote}{%
  \AuthorInstitution, \AuthorAddress,
  \href{mailto:\AuthorEmail}{\AuthorEmail},
  {\large\orcidlink{\AuthorORCID}}\,%
  \href{https://orcid.org/\AuthorORCID}{\AuthorORCID}%
}

\newcommand{\PaperVersion}{Preprint}
\newcommand{\PaperDate}{\today}

\author{}
\date{\PaperDate}

\lstdefinestyle{rcode}{
  language=R,
  basicstyle=\ttfamily\footnotesize,
  backgroundcolor=\color{codebg},
  frame=single,
  framerule=0pt,
  framesep=0.75em,
  breaklines=true,
  breakatwhitespace=true,
  columns=fullflexible,
  keepspaces=true,
  showstringspaces=false,
  keywordstyle={},
  commentstyle={},
  stringstyle={},
  numbers=none,
  captionpos=b,
  xleftmargin=0.5em,
  xrightmargin=0.5em,
  aboveskip=0.8em,
  belowskip=0.8em
}

\begin{document}

{
\setlength{\parindent}{0pt}
\thispagestyle{plain}
{\fontsize{18}{20}\selectfont\raggedright 
\maketitle  

}

{
   \vskip 13.5pt\relax \normalsize\fontsize{11}{12} 
\textbf{\authorfont \AuthorName\footnote{\AuthorFootnote}} \hskip 15pt \emph{\small }   

}

}

\noindent \singlespacing 

\begin{abstract}

    \hbox{\vrule height .2pt width 39.14pc}

    \vskip 8.5pt 

\noindent \textbf{Abstract:} This paper investigates the performance of Echo State Networks (ESNs)
for univariate forecasting of monthly and quarterly time series from the
M4 Forecasting Competition dataset. We evaluate whether a simple
first-order autoregressive ESN can serve as a competitive alternative to
widely used forecasting methods. The study uses a two-stage design: a
\emph{Parameter} dataset is used to analyze ESN model configurations
over leakage rate, spectral radius, reservoir size, and regularization
selection, while a disjoint \emph{Forecast} dataset is reserved for
out-of-sample benchmarking. Forecast accuracy is measured using mean
absolute scaled error (MASE), symmetric mean absolute percentage error
(sMAPE), and the overall weighted average (OWA). The ESN is compared
with simple benchmarks, statistical models including autoregressive
integrated moving average (ARIMA), exponential smoothing state space
(ETS), the Theta method, and TBATS, as well as multilayer perceptron 
(MLP), recurrent neural network (RNN), and hybrid exponential 
smoothing--RNN (ES-RNN) benchmarks. The model-configuration analysis reveals frequency-specific
patterns: monthly series tend to favor moderately persistent reservoirs,
whereas quarterly series favor more contractive dynamics; across both
frequencies, high leakage rates are generally preferred. In the final
benchmark, the ESN performs on par with ARIMA and TBATS under mean MASE
for monthly data and achieves the second-lowest mean MASE for quarterly
data. ES-RNN provides the best aggregate accuracy for both frequencies.
Overall, the results indicate that a simple autoregressive ESN can
provide competitive forecast accuracy on the considered filtered M4
subsets, particularly under MASE, while requiring low training and
forecasting time once the ESN configuration has been fixed.

 \par

    \hbox{\vrule height .2pt width 39.14pc}

\end{abstract}

\vskip -8.5pt


\section{Introduction} \label{sec:introduction}

\extrafootertext{\PaperVersion{} (\PaperDate)} Large-scale forecasting
applications often require many time series to be processed automatically
under limited computational resources. In such settings, forecasting
methods must not only provide accurate predictions, but also be scalable,
easy to train, and suitable for heterogeneous time series characteristics.
Statistical forecasting methods such as autoregressive integrated moving
average (ARIMA), exponential smoothing state space (ETS), the Theta
method, and TBATS remain strong benchmarks because they are well
established, widely implemented, and often competitive in empirical
forecasting studies
\citep{Hyndman2008a, Assimakopoulos2000, DeLivera2011}. Fully trained
recurrent neural networks (RNNs), by contrast, can capture nonlinear
temporal dependencies but typically require more complex training
procedures, larger computational resources, and careful tuning.

Echo State Networks (ESNs), and reservoir computing (RC) more broadly,
offer an intermediate approach between these two model classes. ESNs
retain recurrent nonlinear dynamics, but only the linear readout layer
is trained, while the recurrent reservoir remains fixed after random
initialization
\citep{Jaeger2001, Lukosevicius2009, Lukosevicius2012}. This makes ESNs
attractive for automated forecasting settings in which many time series
must be modeled with limited manual intervention. The present study
therefore examines whether a simple first-order autoregressive ESN can
provide competitive forecasting performance on monthly and quarterly
time series from the M4 Forecasting Competition dataset
\citep{Makridakis2020}.

Although ESNs and RC have been studied in a variety of forecasting
contexts, less systematic evidence is available on how simple
autoregressive ESNs perform across large heterogeneous benchmark
collections of real-world time series under controlled
model-configuration searches and standardized forecast-evaluation
procedures. Existing work often focuses on specific applications,
selected benchmark problems, alternative reservoir topologies, or more
complex ESN and RC variants \citep{Viehweg2025}, while providing limited
guidance on how basic ESN design choices such as leakage rate, spectral
radius, and reservoir size affect forecasting performance across
different data frequencies and time series lengths. This issue is particularly 
relevant when only relatively short time
series are available, as is common, for instance, in business
forecasting settings where structural breaks, organizational changes, or
new product introductions can reduce the usefulness of long historical
records.

Time series forecasting competitions such as M3
\citep{Makridakis2000} and M4 \citep{Makridakis2020} have highlighted
the importance of evaluating forecasting methods on extensive, diverse,
and realistic datasets using standardized metrics and benchmark
procedures. The M4 dataset, in particular, comprises 100,000 series from
multiple application domains and frequencies, providing a stringent
testbed for assessing forecast accuracy across heterogeneous time
series. While machine learning (ML) and hybrid methods featured
prominently among the top-performing approaches in the M4 competition,
ESNs have not yet been systematically investigated in this context. It
therefore remains an open question whether a comparatively simple,
autoregressive ESN can achieve competitive accuracy relative to
well-established statistical benchmarks when applied to monthly and
quarterly time series with realistic history lengths.

This paper addresses this gap by proposing and empirically evaluating an
ESN framework for univariate time series forecasting on a subset of the
M4 dataset. The focus is on monthly and quarterly time series, which are
particularly important for business and economic forecasting, and on
time series with at most 20 years of historical data to reflect
practical constraints encountered in many real-world applications. The
ESN considered here is purely autoregressive, using only the first lag
of the target series as input, and is embedded in a generic
preprocessing pipeline that includes stationarity testing using the 
Kwiatkowski--Phillips--Schmidt--Shin (KPSS) test \citep{Kwiatkowski1992}, 
differencing where required, and scaling to a symmetric interval. 
Forecasts are produced recursively over the standard horizons of the M4 
Forecasting Competition (18 months and 8 quarters), allowing for a 
direct comparison with established benchmark methods.

A key contribution of this study is a comprehensive empirical analysis
of ESN model configurations at scale. The objective is not to introduce
a new ESN architecture or a new training algorithm, but to provide a
controlled assessment of how a first-order autoregressive ESN behaves in
large-scale univariate time series forecasting. We construct two
disjoint random subsets of the M4 data: a \textit{Parameter} dataset
used for systematic analysis of ESN hyperparameters and model
configurations, and a \textit{Forecast} dataset reserved for
out-of-sample evaluation. On the \textit{Parameter} dataset, we perform
an extensive grid search over four ESN design dimensions, i.e., the
leakage rate, spectral radius, reservoir size, and choice of information
criterion used to select the ridge penalty. This two-stage procedure
allows us to examine how ESN design choices affect forecasting accuracy
across thousands of heterogeneous time series and to identify
frequency-specific configurations that are subsequently evaluated on
unseen series.

The predictive performance of the selected ESN configurations is then
evaluated on the \textit{Forecast} dataset and compared against a set of
widely used benchmark methods. These include simple approaches (naive, 
Naive2, drift, seasonal naive, and mean forecasts), statistical models
(ARIMA, ETS, Theta, and TBATS), and neural network (NN) benchmarks, including 
the multilayer perceptron (MLP), RNN, and the winning exponential 
smoothing--recurrent neural network (ES-RNN) model from the M4 
Forecasting Competition. Forecast accuracy 
is assessed using three metrics from the forecasting literature: the 
symmetric mean absolute percentage error (sMAPE), the 
mean absolute scaled error (MASE), and the overall weighted average
(OWA) \citep{Hyndman2006a, Makridakis2020}. In addition to accuracy, we
report computational runtime for the locally estimated methods,
providing a joint view of the trade-off between predictive performance
and efficiency that is central to large-scale, automated forecasting.

The empirical results indicate that the proposed ESN is competitive
with established statistical benchmarks, although its relative
performance depends on the data frequency and accuracy metric
considered. For monthly series, the selected ESN configuration performs
on par with ARIMA and TBATS in terms of mean MASE. For quarterly series,
the ESN achieves the second-lowest mean MASE and ranks third according
to OWA. ES-RNN provides the best aggregate forecast accuracy for both
frequencies, while the standalone MLP and RNN benchmarks perform less
favorably. Pairwise statistical tests show that the ESN is generally
statistically indistinguishable from the strongest statistical
benchmarks. The runtime comparison further shows that the selected ESN
configurations can be trained and applied efficiently once the model
configuration has been fixed. The analysis of hyperparameters and model
configurations reveals systematic frequency-specific patterns: both
frequencies favor high leakage rates, but the preferred spectral radius
and reservoir size differ between monthly and quarterly data,
underlining the importance of adapting reservoir dynamics to data
frequency and time series length.

Overall, this study contributes to the empirical forecasting literature
in three ways:
\begin{enumerate}
    \item It evaluates a simple first-order autoregressive ESN framework
    for univariate time series forecasting in an automated, large-scale
    setting.
    \item It provides an empirical analysis of ESN hyperparameters and
    model configurations on a heterogeneous benchmark dataset, yielding
    practical guidance for future applications.
    \item It compares the selected ESN configurations with simple,
    statistical, and NN forecasting methods in terms of forecast
    accuracy and computational runtime, thereby clarifying the conditions
    under which this simple ESN specification constitutes a competitive
    empirical alternative.
\end{enumerate}

The remainder of the paper is structured as follows. Section
\ref{sec:data} describes the dataset and sampling strategy. Section
\ref{sec:esn} details the ESN model, preprocessing steps, and training
procedure. Section \ref{sec:empirical-analysis} presents the empirical
application, including accuracy metrics, the hyperparameter sweep,
random initialization analysis, and benchmark comparison. The paper
closes with a summary and concluding remarks. The Appendix provides
additional empirical results and documents the computational workflow,
software environment, replication code, and computing hardware.

\section{Data} \label{sec:data}

The dataset used in the empirical application is taken from the
well-known M4 Forecasting Competition \citep{Makridakis2020}. The
original M4 dataset consists of 100,000 time series, subdivided into six
data frequencies (hourly, daily, weekly, monthly, quarterly, and yearly)
and six application fields (Micro, Macro, Industry, Finance,
Demographic, and Other). The time series were randomly selected from a
database called ForeDeCk, compiled at the National Technical University
of Athens (NTUA). The database contains 900,000 time series from
multiple publicly available sources. The focus of ForeDeCk is business
forecasting; therefore, the time series come from domains such as
industries, services, tourism, imports and exports, demographics,
education, labor and wage, government, households, bonds, stocks,
insurance, loans, real estate, transportation, and natural resources and
environment \citep{Spiliotis2020}.

In preparation for the M4 Forecasting Competition, the dataset was
scaled to avoid negative values and values lower than 10 to circumvent
problems when calculating forecast error metrics. Scaling the data by
adding a constant value to the time series ensures that their minimum
value equals 10. Additionally, any information pointing to the original
time series was removed to guarantee the objectivity of the results.
Low-volume and intermittent time series data were not considered to
avoid methodological difficulties with close-to-zero and zero values.
Therefore, the dataset primarily tests the accuracy of different
forecasting methods on continuous time series \citep{Makridakis2020}.

The number of time series for each frequency and application field was
determined mainly based on the importance for organizations in terms of
operational, tactical, and strategic planning and forecasting. For
example, creating monthly and quarterly forecasts is more often required
in business and economic forecasting than hourly or yearly forecasts.
Moreover, micro and financial data are more frequently used than
demographic data \citep{Makridakis2020, Spiliotis2020}.

In order to evaluate the forecast accuracy of the proposed ESN, only
monthly and quarterly time series across all application fields were
considered. Some of the time series in the M4 dataset are very long,
with more than 200 years of history. While some applications, such as
macroeconomic forecasting, have access to extensive historical data,
other applications, like business planning and forecasting, typically
rely on a few years of historical data, if any. For example, changes in
business operations, such as structural or organizational changes within
a company, can render older data less applicable. This distinction
ensures that our approach remains relevant and applicable to a broad
range of real-world scenarios, where only shorter time series are
available. Therefore, only time series with at most 20 years of history
were retained, corresponding to no more than 240 monthly or 80 quarterly
observations.

Given this restriction, 28,820 monthly and 10,950 quarterly time series remain from the available 48,000 monthly and 24,000 quarterly series. From this filtered collection, two disjoint subsets are constructed by random sampling. The first subset, referred to as the \textit{Parameter} dataset, contains 2,400 monthly and 1,200 quarterly time series and is used for the systematic analysis of ESN hyperparameters and model configurations, including leakage rate, spectral radius, reservoir size, and the information criterion used for regularization selection. The second subset, referred to as the \textit{Forecast} dataset, contains the same number of time series and is sampled independently after excluding all series used in the \textit{Parameter} dataset. It is reserved for out-of-sample evaluation and benchmarking against standard forecasting methods.

To ensure reproducibility, the random sampling was conducted using a fixed random seed after applying the frequency and maximum-history filters. The identifiers of the selected M4 series are made available with the replication code.

The \textit{Parameter} dataset is used exclusively to evaluate ESN
hyperparameter configurations and to identify the best-performing
specification separately for monthly and quarterly data based on average
forecast accuracy across all series in that dataset. Once selected,
these frequency-specific ESN configurations are fixed and carried
forward unchanged to the disjoint \textit{Forecast} dataset, where they
are evaluated against the benchmark methods. This two-stage approach
helps to mitigate overfitting and provides an out-of-sample
assessment of the selected ESN configurations. By structuring
the analysis in this way, the study establishes a transparent framework
for assessing the forecasting potential of ESNs. Based on the M4
Forecasting Competition, the forecast horizon for the monthly data is
1.5 years (\(h = 18\)), and for the quarterly data, two years
(\(h = 8\)). Table \ref{tab:dataset} in the Appendix summarizes the
absolute and relative frequencies of both sampled datasets according to
the application field.

To characterize the sampled datasets and to assess how they differ from the broader M4 collection, an exploratory data analysis is conducted for key time series characteristics. In particular, we compare the distributions of series length, seasonal strength, and trend strength, since these characteristics are important for many forecasting methods. Some methods explicitly model level, trend, and seasonality, whereas others capture only a subset of these components. The number of observations is also relevant because more flexible, data-driven forecasting methods typically require more historical data than simple benchmark methods.

\begin{figure}[hbt!]
\centering
\includegraphics[width=\linewidth]{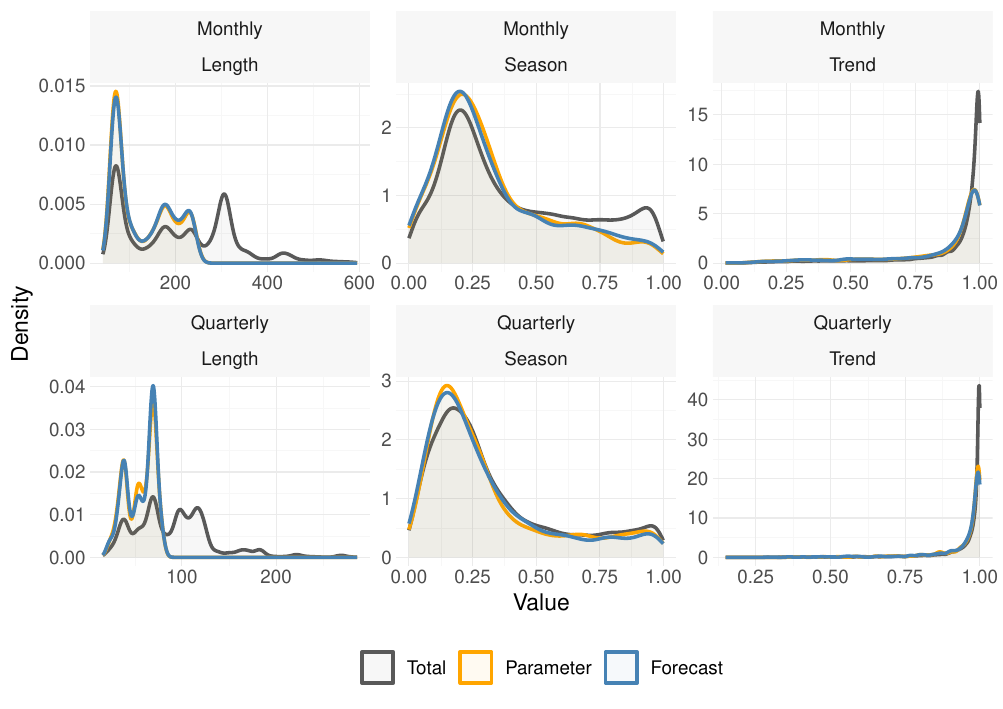}
\caption[Distributions of time series characteristics]{Density plots illustrating the distributions of time series characteristics for the monthly (top row) and quarterly (bottom row) data. The left panels show the number of observations (series length), the middle panels display the strength of seasonality, and the right panels depict the strength of trend. The black lines represent the total M4 dataset, while the orange and blue lines correspond to the randomly sampled \textit{Parameter} and \textit{Forecast} datasets (Data source: M4 Forecasting Competition).}
\label{fig:data-summary}
\end{figure}

Figure \ref{fig:data-summary} shows the distributions of the number of
observations, the seasonality strength, and the trend strength for the
monthly and quarterly time series in both sample datasets and the total
dataset for comparison. Details of the calculation for strength of trend
and strength of seasonality are provided in the Appendix. The top row
visualizes the monthly and the bottom row the quarterly data. The left
panel shows the number of observations (i.e., the time series length),
the middle panel shows the strength of the seasonality and the right
panel shows the strength of trend. The total dataset is colored in
black, while the dataset \textit{Parameter} is plotted in orange and
\textit{Forecast} is shown in blue. For visual clarity, the x-axis in
Figure \ref{fig:data-summary} (left panel) is truncated at 600
observations for the monthly data and 300 observations for the quarterly
data when plotting the full M4 reference distributions. This affects
only the graphical display: the sampled \textit{Parameter} and
\textit{Forecast} datasets remain restricted to series with at most 240
monthly or 80 quarterly observations, and all descriptive analyses are
based on the complete filtered samples.

For the monthly series, the full M4 dataset shows an average series
length of about 216 observations, with a broad range from 42 to nearly
2,800 observations, reflecting the heterogeneity of the underlying time
series sources. The two sampled subsets, however, are shorter on
average, with mean lengths of around 128 observations and a maximum of
240, reflecting the imposed maximum-history filter and resulting in shorter,
more homogeneous datasets suitable for model experimentation. In terms of seasonality, the average seasonal strength
in the total dataset (mean 0.415) is somewhat higher than in the sampled
subsets (means around 0.34), suggesting that the selected series may
exhibit slightly weaker seasonal components on average. The trend
strength remains consistently high across all datasets, with mean values
around 0.85--0.89 and median values close to 0.95, indicating that most
monthly time series in the M4 dataset display a pronounced trend. The
small differences between the total and sampled datasets confirm that
the sampling preserved the essential trend characteristics of the
original collection.

\begin{figure}[hbt!]
\centering
\includegraphics[width=\linewidth]{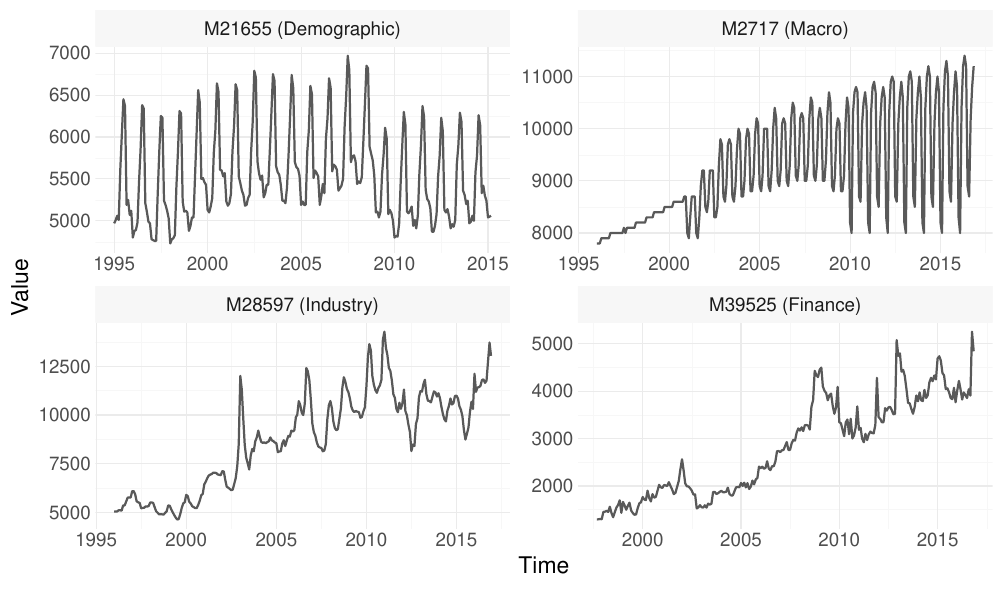}
\caption[Example monthly time series]{Exemplary monthly time series with varying characteristics. The four time series are taken from the randomly sampled \textit{Parameter} dataset (Data source: M4 Forecasting Competition).}
\label{fig:data-sample}
\end{figure}

For the quarterly series, a similar pattern emerges. The full M4 dataset
contains an average of about 92 observations per series, while the
sampled subsets have a shorter mean length of approximately 56
observations, again reflecting the imposed maximum-history filter.
The distribution of seasonal strength shows comparable behavior across
datasets, with the sampled subsets having slightly lower mean values
(around 0.30--0.31) than the total dataset (0.34), indicating a minor
reduction in seasonality. The trend strength is consistently very high
across all datasets, with mean values exceeding 0.94 and median values
above 0.98, confirming that most quarterly series in the M4 dataset
exhibit strong trending behavior. Tables \ref{tab:descriptive-monthly}
and \ref{tab:descriptive-quarterly} in the Appendix provide a numeric
summary of the time series characteristics. Figure \ref{fig:data-sample}
shows four exemplary monthly time series with varying characteristics.

The sampled datasets preserve several important characteristics of the filtered monthly and quarterly M4 series, particularly the broad distributions of trend strength and series length within the imposed maximum-history constraint. At the same time, the filtering procedure changes the composition of the data: the sampled series are shorter by construction and exhibit slightly weaker seasonality on average than the full monthly and quarterly M4 collections. Consequently, the empirical conclusions should be interpreted as applying to this filtered forecasting setting rather than to the full M4 benchmark.

\section{Echo State Networks} \label{sec:esn}

Forecasting time series in our setting requires learning the
relationship between past observations and future values. Since many
time series exhibit nonlinear dynamics, purely linear models are often
too restrictive to capture these dependencies adequately. A common
approach in this context is to map the input into a high-dimensional
nonlinear feature space and then estimate the forecast by means of a
simple linear readout, such as regression \citep{Lukosevicius2009}.

ESNs share the general architecture of neural networks but differ
fundamentally in training. ESNs separate the nonlinear temporal
processing, handled by a fixed, recurrent reservoir, from the readout
layer, which is typically linear and trained via supervised learning.
This separation contrasts with traditional RNNs, where all weights are
jointly optimized \citep{Lukosevicius2009}.

An ESN consists of an input layer, a large randomly initialized
reservoir, and an output layer. Only the output weights
\(\mathbf{W}^\mathrm{out}\) are trained, whereas the input weight matrix
\(\mathbf{W}^\mathrm{in}\) and the reservoir weight matrix
\(\mathbf{W}\) remain fixed \citep{Lukosevicius2012}. The reservoir's
dynamically evolving internal states provide a rich basis from which the
output is linearly reconstructed.

For univariate forecasting, the output layer contains a single output
variable, \(y_t\). In the autoregressive ESN considered here, the input
is simply the previous observation, \(u_t = y_{t-1}\), analogous to a
first-order autoregressive model. Owing to this autoregressive setup,
ESNs naturally support recursive forecasting (``closed-loop''): the
one-step ahead forecast \(\hat{y}_{T+1 \mid T}\) is fed back as input to
generate the two-step ahead forecast \(\hat{y}_{T+2 \mid T}\), and so
on. The process continues until the desired forecast horizon \(h\) is
reached.

This input specification is intentionally restrictive. It keeps the ESN
architecture comparable across all time series and isolates the effect of the
reservoir dynamics, but it does not explicitly provide seasonal lagged
inputs such as \(y_{t-12}\) for monthly data or \(y_{t-4}\) for
quarterly data. Seasonal information can therefore only enter indirectly
through the reservoir state and the recursive forecast dynamics.
Consequently, the results should be interpreted as the performance of a
simple first-order autoregressive ESN rather than as the best achievable
performance of ESNs with richer input structures.

As with most neural networks, input data must be preprocessed before
training and forecasting. Many series in our dataset are nonstationary
due to trend, seasonality, or heteroscedasticity, and trending behavior
is particularly problematic because the reservoir cannot effectively
handle such nonstationarity. We therefore apply a
Kwiatkowski--Phillips--Schmidt--Shin (KPSS) test \citep{Kwiatkowski1992}
to the training portion of each series to assess whether the output
variable \(y_t\) is level-stationary. The KPSS null hypothesis is
stationarity around a constant level. In the empirical application, the
test is evaluated at the 5\% significance level. If the null hypothesis
of level stationarity is rejected, a single first difference is applied
to stabilize the mean and reduce nonseasonal trend components
\citep{Hyndman2021}. Seasonal differencing is not used in the baseline
specification.

After detrending, the data are scaled to the interval \([-0.5,0.5]\).
Scaling is important because ESNs are sensitive to the magnitude of the
inputs, and keeping the data within a controlled range improves
numerical stability and reservoir dynamics. Empirically, scaling to the
symmetric range \([-0.5, 0.5]\) performed well, though other
normalization methods (e.g., min--max or z-score standardization) are
also suitable.

Scaling parameters are estimated exclusively from the transformed
training data and are then applied during recursive forecasting to avoid
information leakage. After forecasts are generated on the transformed
scale, the inverse scaling transformation is applied first, followed by
reconstruction of the original scale by reversing the differencing
operation using the last available training observation.

In an ESN, the hidden layer (reservoir) is represented by
\(\mathbf{x}_t = (x_{t,1}, x_{t,2}, \ldots, x_{t,N_x})^\top\), where
\(x_{t,n}\) denotes the internal state of reservoir unit \(n\) at time
\(t\), and \(N_x\) is the total number of reservoir units (reservoir
size). Given a single input \(u_t = y_{t-1}\), the reservoir dynamics
follow the standard update equation

\begin{equation}
\tilde{\mathbf{x}}_{t} = \tanh\!\left(\mathbf{W}^{\mathrm{in}} u_{t} + \mathbf{W}\mathbf{x}_{t-1}\right),
\label{eq:reservoir}
\end{equation}

for \(t = 1,2,\dots,T\), with initial condition \(\mathbf{x}_0 = \mathbf{0}\). The matrix \(\mathbf{W}^{\mathrm{in}} \in \mathbb{R}^{N_x \times 1}\) denotes the input weight matrix, while \(\mathbf{W} \in \mathbb{R}^{N_x \times N_x}\) denotes the final
reservoir weight matrix used in the state update. The nonlinear transformation in (\ref{eq:reservoir}) is applied element-wise via the hyperbolic tangent activation function, which is the most common choice in ESN applications \citep{Lukosevicius2012}.

The input weight matrix \(\mathbf{W}^{\mathrm{in}}\) is typically initialized with random values, commonly drawn from a uniform distribution on \([-0.5,0.5]\), although other distributions such as the normal distribution are also possible. In the present study, input scaling is held fixed through this initialization range, which reduces the dimensionality of the hyperparameters and model configurations. Since input scaling can affect ESN performance, the results should be interpreted conditional on this fixed choice for input scaling. For the reservoir, let \(\mathbf{W}_0\) denote the initial random weight matrix. In the present application, \(\mathbf{W}_0\) is constructed to be sparse, with a density of 50\%, meaning that half of its entries are randomly drawn from a uniform distribution on \([-0.5,0.5]\) and the remaining half are set to zero. The reservoir density is held fixed at 50\% to limit the dimensionality of the already extensive hyperparameter sweep and to isolate the effects of leakage rate, spectral radius, reservoir size, and regularization selection. The value is therefore used as a pragmatic baseline rather than as a theoretically or empirically optimal density. Since reservoir sparsity can interact with the spectral radius, the echo state property, and the rate of memory decay, the empirical results should be
interpreted as conditional on this fixed density. Substantially sparser reservoirs are also common in ESN applications, and a systematic sensitivity analysis of reservoir sparsity remains a direction for future research.

A crucial ESN hyperparameter is the spectral radius of the reservoir
matrix. Let \(\rho(\mathbf{W}_0)\) denote the spectral radius of the
initial reservoir matrix \(\mathbf{W}_0\), defined as its largest
absolute eigenvalue. We first normalize \(\mathbf{W}_0\) according to
\(\mathbf{W}_1 = \mathbf{W}_0 / \rho(\mathbf{W}_0)\) so that
\(\mathbf{W}_1\) has unit spectral radius. The final reservoir weight
matrix is then obtained as \(\mathbf{W} = \rho \mathbf{W}_1\), where
\(\rho\) is the target spectral radius used as a hyperparameter. In this
way, \(\rho\) controls the effective scale of the recurrent weight
matrix and thereby influences the reservoir dynamics.

The spectral radius is commonly used as a heuristic control parameter
for the memory and stability of the reservoir. Smaller values usually
lead to more contractive dynamics and faster forgetting, whereas larger
values allow past inputs to influence the reservoir state for longer.
A related concept is the echo state property (ESP), which requires that
the influence of past inputs and initial conditions fades over time so
that the current reservoir state is primarily driven by the recent input
history. The spectral radius can influence this behavior, but the ESP is
not determined by the spectral radius alone; it also depends on leakage,
input scaling, the activation function, and the realized reservoir
weights. Therefore, values close to or above one should be interpreted
as exploratory settings near the boundary of stable reservoir dynamics
rather than as formal guarantees of the ESP \citep{Lukosevicius2012}.

In the empirical application, the reservoir size is chosen according to
the rule of thumb
\(N_x = \min \bigl( \lfloor \tau T \rfloor,\, 200 \bigr)\), with
\(\tau\) as the so-called reservoir scaling parameter and \(T\) denoting
the sample size. Thus, the number of reservoir units is proportional to
the available observations, rounded down using the floor operator. The
upper bound of 200 prevents excessively large reservoirs for long time
series. Although this rule is linear in \(T\), nonlinear alternatives
could also be considered. This rule of thumb is a pragmatic design
choice adopted for the empirical application and is not derived from a
formal theoretical result. Its purpose is to provide a simple and
consistent way to adapt reservoir size to time series length while
keeping the overall model complexity computationally manageable across a
large number of series.

For a leaky integrator ESN, a leakage rate \(\alpha \in (0,1]\) is
introduced:

\begin{equation}
\mathbf{x}_t = (1 - \alpha)\,\mathbf{x}_{t-1} + \alpha \tilde{\mathbf{x}}_t 
\label{eq:leaky-reservoir}
\end{equation}

This update blends the previous internal state with the newly computed
nonlinear state, introducing a controllable form of memory analogous to
exponential smoothing. Setting \(\alpha = 1\) recovers the standard
(non-leaky) ESN, while \(0 < \alpha < 1\) produces a weighted average of
\(\mathbf{x}_{t-1}\) and the \(\tanh(\cdot)\) term. Smaller values of
\(\alpha\) slow the reservoir dynamics and enable the network to capture
longer-term dependencies \citep{Lukosevicius2012}.

The internal states are computed according to equations
\ref{eq:reservoir} and \ref{eq:leaky-reservoir} and, together with an
intercept term, stacked over time to form the design matrix
\(\mathbf{X}\). Accordingly, each row of \(\mathbf{X}\) is of the form
\((1, \mathbf{x}_t^\top)\), where the first column contains ones and the
remaining columns contain the reservoir states at time \(t\).

Initializing the reservoir at \(\mathbf{x}_{0}=0\) can lead to an
artificial transient phase during which the reservoir dynamics have not
yet stabilized. It is therefore standard practice to discard an initial
portion of the observations, often referred to as a washout or warm-up
period, before estimating the output weights
\citep{Lukosevicius2012}. Based on preliminary empirical experimentation
conducted as part of the present study, the first
\(\delta = \lfloor 0.05 T \rfloor\) observations are removed from the
design matrix. This proportional rule allows the washout period to scale
with the time series length while retaining 95\% of the observations for
readout estimation. The value should be interpreted as a pragmatic
implementation choice rather than as a theoretically optimal washout
length.

Overfitting is a common challenge in machine learning and particularly
in ESNs, where models may capture both signal and noise. The goal is to
obtain a regularized model with good forecasting performance, which
requires controlling overfitting and reducing prediction error. In the
ESN literature \citep{Jaeger2001, Jaeger2002}, ridge regression is used
for training the linear readout. When many (often highly correlated)
features are present, linear models can suffer from multicollinearity,
which may impair forecast performance; ridge regression mitigates these
issues \citep{Hoerl1970}.

Regularization helps stabilize ESN training by addressing overfitting
and ill-posed optimization. Ridge regression resembles least squares,
but shrinks coefficients toward zero. Given the output vector
\(\mathbf{y}\) and the design matrix \(\mathbf{X}\) of internal states,
the linear model is

\begin{equation}
\mathbf{y} = \mathbf{X} \mathbf{W}^\mathrm{out} + \boldsymbol{\epsilon},
\label{eq:linear-model}
\end{equation}

\noindent where \(\mathbf{y}\) is \(T \times 1\), \(\mathbf{X}\) is
\(T \times (N_x+1)\) and includes an intercept, and
\(\boldsymbol{\epsilon}\) is an error term.

Ridge regression introduces a penalty on coefficient size, analogous to
weight decay in neural networks \citep{Hastie2009}. The estimator is
defined as

\begin{equation}
\mathbf{\widehat{W}}^\mathrm{out} = \left(\mathbf{X}^\top \mathbf{X} + \mathbf{R}_{\lambda}\right)^{-1}\mathbf{X}^\top\mathbf{y},
\label{eq:estimator}
\end{equation}

\noindent where
\(\mathbf{R}_{\lambda} = \operatorname{diag}(0, \lambda, ..., \lambda)\)
ensures that the intercept is unpenalized while all \(N_x\) coefficients
share a common penalty \(\lambda\). Adding \(\mathbf{R}_{\lambda}\)
resolves rank-deficiency in \(\mathbf{X}^\top \mathbf{X}\) and thereby
stabilizes the matrix inversion \citep{Hastie2009}.

Selecting an appropriate \(\lambda\) requires a transparent and
computationally feasible tuning strategy. In this study, information
criteria are used as pragmatic selection rules for the ridge penalty
within each fixed ESN configuration. Because the reservoir induces a
nonlinear transformation of the original time series, these criteria
should not be interpreted as exact likelihood-based model selection
criteria for the full ESN process. Rather, they provide a consistent way
to compare ridge readouts fitted to the same reservoir-state
representation. Model complexity is measured through the effective
degrees-of-freedom, defined via the trace of the ridge hat matrix
\citep{Hastie2009}:

\begin{equation}
df_{\lambda}
=
\operatorname{tr}\left(\mathbf{H}_{\lambda}\right)
=
\operatorname{tr}\left(
\mathbf{X}
\left(
\mathbf{X}^{\top}\mathbf{X} + \mathbf{R}_{\lambda}
\right)^{-1}
\mathbf{X}^{\top}
\right).
\label{eq:dof}
\end{equation}

Several hyperparameter search methods exist, but grid search and random
search remain widely used \citep{Bergstra2012}. Grid search evaluates
all points in a predefined grid, while random search samples values from
a bounded domain, often uniformly. \citet{Bergstra2012} show that random
search is typically more efficient, often finding equal or better models
with far fewer evaluations.

This paper employs random search over \(\lambda \in [10^{-4}, 2]\), an interval chosen through preliminary experimentation. Specifically, for each fixed ESN configuration defined by \(\alpha\), \(\rho\), and \(\tau\), we draw \(K = 2N_x\) candidate values of \(\lambda\) uniformly, thereby scaling the search size with the reservoir dimension. For each candidate, we fit the linear readout and compute the corresponding information criterion, which is used solely to select the ridge regularization parameter within the given ESN specification. The candidate with the lowest criterion value is then selected for forecasting. Time series cross-validation (rolling origin forecasting) could provide an alternative basis for choosing \(\lambda\), but would substantially increase the computational cost given the number of series and ESN configurations considered. Details on the information criteria are provided in Appendix \hyperref[sec:appendix-hyperparameter]{A.2}.

\section{Empirical analysis} \label{sec:empirical-analysis}

\subsection{Accuracy metrics} \label{sec:empirical-metrics}

Many error metrics are available in the forecasting literature to
evaluate the accuracy of forecasting methods. Here, the out-of-sample
forecasting performance is measured by the symmetric mean absolute
percentage error (sMAPE), the mean absolute scaled error (MASE), and the
overall weighted average (OWA). The sMAPE and MASE are defined as

\begin{equation}
sMAPE = \frac{2}{h} \sum\nolimits_{t=T+1}^{T+h}
\frac{\left| y_{t} - \hat{y}_{t} \right|}
{\left| y_{t} \right| + \left| \hat{y}_{t} \right|} 100\%,
\label{eq:smape}
\end{equation}

\begin{equation}
MASE = \frac{1}{h}
\frac{\sum\nolimits_{t=T+1}^{T+h}
\left| y_{t} - \hat{y}_{t} \right|}
{\frac{1}{T-m}\sum\nolimits_{t=m+1}^{T}
\left| y_{t} - y_{t-m} \right|},
\label{eq:mase}
\end{equation}

\noindent where \(y_t\) is the actual value and \(\hat{y}_t\) is the
estimated point forecast for a given period, \(h\) is the forecast
horizon, \(T\) is the number of observations available in-sample, and
\(m\) is the frequency of the time series, i.e., \(m = 12\) for monthly
and \(m = 4\) for quarterly data
\citep{Hyndman2021, Hyndman2006a, Makridakis2020}.

The sMAPE calculates the absolute difference between the actual value
and the forecast (the numerator in equation (\ref{eq:smape})). It
divides by half the sum of the absolute value of the actuals and the
absolute value of the forecasts (the denominator in equation
(\ref{eq:smape})). This value is then summed for all points in time and
divided by the forecast horizon.

In contrast to other percentage metrics, the sMAPE treats positive and
negative forecast errors equally (symmetric). Percentage error metrics
cannot be used when there are zero or close-to-zero values because there
would be a division by zero, or the values would tend to infinity. Thus,
the metric can be unstable. In the present application, however, this
concern is substantially reduced because the original M4 dataset was
specifically prepared to exclude negative, zero, and near-zero values.
The sMAPE is still a helpful error metric, as it is easy to interpret.

The MASE is one of the most widely used forecast error metrics in
contemporary forecasting literature. It overcomes the issues of the
other metrics mentioned above and is more suitable from a research
perspective. The idea of MASE is to scale the out-of-sample mean
absolute error (MAE) of the forecasting method at hand (the numerator in
equation (\ref{eq:mase})) with the in-sample MAE of some benchmark
method (the denominator in equation (\ref{eq:mase})). In the original
publication of Hyndman and Köhler \citep{Hyndman2006a}, the naive
forecast method (\(m = 1\)) is proposed as the benchmark. However,
benchmark methods such as the seasonal naive approach (\(m = 12\) or
\(m = 4\)) provide a more appropriate benchmark for seasonal time
series \citep{Makridakis2020}.

The MASE has some appealing properties when comparing the accuracy of
different methods and time series. For example, the MASE is
scale-invariant, meaning that it is independent of the scale of the
data. Hence, it allows for comparisons of forecasts across different
scales (e.g., revenue in millions EUR vs.~revenue in thousands EUR).
There are no problems with zero or close-to-zero values because there is
no division by the actual values, and the MASE is symmetric in that
positive and negative forecast errors are treated equally. Finally, MASE
is smaller than one if the examined forecast method performs better than
the benchmark, and MASE is larger than one when the benchmark method
performs better \citep{Hyndman2021, Hyndman2006a, Makridakis2020}.

To provide a combined measure of forecast accuracy, the M4 Forecasting
Competition uses the OWA, which averages the relative sMAPE and MASE of
a forecasting method compared with the Naive2 benchmark
\citep{Makridakis2020}. For model \(j\), it is defined as

\begin{equation}
OWA_j =
\frac{1}{2}
\left(
\frac{\overline{sMAPE}_j}
{\overline{sMAPE}_{\mathrm{NAIVE2}}}
+
\frac{\overline{MASE}_j}
{\overline{MASE}_{\mathrm{NAIVE2}}}
\right),
\label{eq:owa}
\end{equation}

\noindent where \(\overline{sMAPE}_j\) and
\(\overline{MASE}_j\) denote the mean accuracy values of model \(j\)
across all time series. By construction, Naive2 has an OWA of one.
Values below one indicate that a method performs better overall than
Naive2, while values above one indicate worse overall performance. The
OWA therefore provides a single ranking measure that gives equal weight
to the relative sMAPE and MASE results.

\subsection{Hyperparameter sweep} \label{sec:hyperparameter-sweep}

To systematically explore the effect of hyperparameter configurations,
we perform a comprehensive hyperparameter sweep based on an exhaustive
grid search. The grid spans four core ESN parameters: the leakage rate
\(\alpha\), the spectral radius \(\rho\), the reservoir scaling factor
\(\tau\) and the information criterion. The leakage rate is varied in
equidistant steps from 0.1 to 1.0, allowing the model to interpolate
between slowly adapting reservoirs with long memory and rapidly updating
reservoirs that prioritize short-term dynamics. The spectral radius is
swept from 0.2 to 1.2 in increments of 0.1, covering both strictly
contractive regimes as well as values slightly above unity, which are
known to place the reservoir near the edge of stability and can
empirically enhance predictive performance in certain settings. The
reservoir size is explored via the scaling parameter
\(\tau \in \{0.2, 0.4, 0.6\}\), which allows the capacity of the model
to adjust to the length of each time series while maintaining a fixed
upper bound to avoid excessive complexity. Each configuration was
further evaluated using information criteria including Akaike
Information Criterion (AIC) \citep{Akaike1974}, corrected Akaike
Information Criterion (AICc) \citep{Sugiura1978, Hurvich1989}, Bayesian
Information Criterion (BIC) \citep{Schwarz1978}, and Hannan-Quinn
Criterion (HQC) \citep{Hannan1979} to account for model parsimony
relative to goodness of fit. These criteria provide a complementary
perspective by penalizing overly complex reservoir configurations,
thereby supporting a more balanced model selection beyond pure
predictive performance.

The resulting grid comprises 1,320 distinct hyperparameter
configurations (\(10 \times 11 \times 3 \times 4\)), each of which is
evaluated independently for every time series. This yields a total of
4,752,000 ESN fits across both frequencies (1,320 configurations across
2,400 monthly and 1,200 quarterly series), which remains computationally
feasible due to the closed-form training of the linear readout. The
chosen search space is deliberately compact yet expressive: it covers
qualitatively distinct dynamical regimes while avoiding unnecessarily
fine-grained sampling that would not lead to practically meaningful
differences in model behavior. By evaluating each configuration independently on each time series, the
procedure accommodates the heterogeneous temporal characteristics within
the datasets and supports the selection of a frequency-specific
hyperparameter configuration based on aggregate forecast accuracy.

\begin{figure}[hbt!]
\centering
\includegraphics[width=\linewidth]{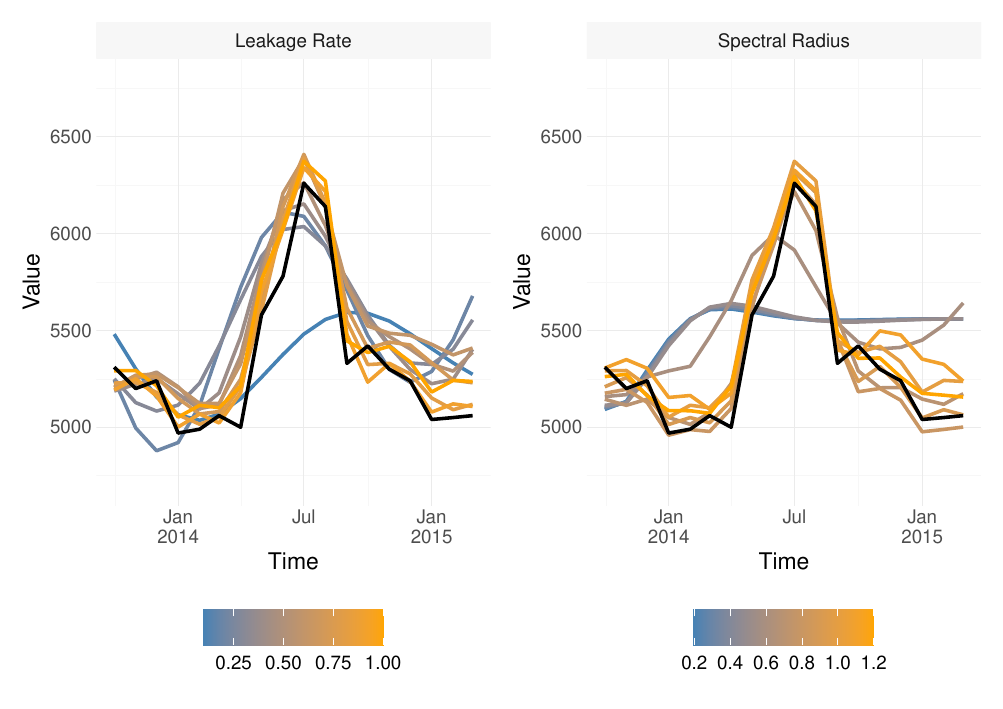}
\caption[Forecasts under varying ESN hyperparameters]{Forecasts for the example monthly series M21655 under varying hyperparameter settings. Left: forecasts for leakage rates $\alpha \in \{0.1, 0.2, ..., 0.9, 1.0\}$. Right: forecasts for spectral radii $\rho \in \{0.2, 0.3, ..., 1.1, 1.2\}$. Actual observations (holdout data) are shown in black. Colored forecast lines transition from blue (low values) to orange (high values), illustrating how increasing $\alpha$ and $\rho$ alters forecast smoothness, reactivity, and temporal lag.}
\label{fig:fcst-pars}
\end{figure}

To illustrate the qualitative effect of individual hyperparameters on
the forecast dynamics, Figure \ref{fig:fcst-pars} presents the results
for the example series M21655 from the monthly dataset. The left panel
shows forecasts obtained for different leakage rates \(\alpha\) (ranging
from 0.1 to 1.0), while the right panel visualizes forecasts for varying
spectral radii \(\rho\) (ranging from 0.2 to 1.2). The actual holdout
observations are plotted in black, and the forecasts are color-coded
from blue to orange to reflect increasing hyperparameter values. As
expected, low leakage values lead to smoother and more inert forecast
trajectories, whereas higher leakage rates produce more reactive but
also more volatile forecasts. A similar pattern is observed for the
spectral radius: small \(\rho\) values result in strongly damped
dynamics, while larger values allow the reservoir to retain more
temporal information that amplifies fluctuations. This visual example
highlights how the hyperparameter configuration governs the
responsiveness and memory characteristics of the ESN, motivating the
need for a systematic search across a broad grid rather than relying on
a single default setting.

To complement the visualization-based analysis of individual
hyperparameter effects, Tables \ref{tab:pars-monthly-top} and
\ref{tab:pars-quarterly-top} summarize the 30 best-performing ESN
configurations for the monthly and quarterly datasets, respectively. By
ranking configurations based on mean MASE and reporting both MASE and
sMAPE alongside the corresponding hyperparameters and information
criteria, these tables allow for a structured comparison of recurring
patterns in optimal settings across frequencies.

Table \ref{tab:pars-monthly-top} lists the 30 best-performing
hyperparameter configurations for the monthly dataset, ranked by mean
MASE. Across the top-ranked configurations, a clear pattern emerges:
almost all high-performing models operate with a leakage rate of
\(\alpha = 0.9\) or \(\alpha = 1.0\), indicating that more reactive
reservoirs with limited smoothing tend to be favorable when sufficient
data points are available, as is the case in monthly series of typical
length. Similarly, the spectral radius concentrates around
\(\rho = 0.8\) to \(\rho = 1.0\), placing the reservoir close to the
edge of stability. This confirms the earlier observation from the
visualization example that slightly more expressive and less damped
dynamics improve forecasting accuracy. In terms of reservoir size
scaling, \(\tau = 0.4\) appears most frequently among the top-ranked
configurations, with \(\tau = 0.6\) occurring in some cases but never
dominating. This suggests that moderately sized reservoirs are
sufficient to capture relevant temporal structure, whereas further
increasing the number of units does not consistently translate into
better performance. Across all information criteria, AICc and HQC appear
most frequently among the top configurations, but differences remain
minor, showing that the selection of the information criteria is less
important.

\begin{table}
\centering
\begin{tabular}{rllrrrrrrr}
\toprule
\multirow{2}{*}{\textbf{Rank}} & \multirow{2}{*}{\textbf{Model}} & \multicolumn{4}{c}{\textbf{Model Specification}} & \multicolumn{2}{c}{\textbf{MASE}} & \multicolumn{2}{c}{\textbf{sMAPE [\%]}} \\
\cmidrule(l){3-6} \cmidrule(l){7-8} \cmidrule(l){9-10}
 & & IC & $\alpha$ & $\rho$ & $\tau$ & Mean & Median & Mean & Median \\
\midrule
\rowcolor{light-gray}
1 & ESN-0650 & AICc & 1.0 & 0.9 & 0.4 & \textbf{0.878} & 0.712 & 17.592 & 12.406 \\ 
2 & ESN-1310 & HQC & 1.0 & 0.9 & 0.4 & 0.878 & 0.715 & 17.601 & 12.438 \\ 
3 & ESN-0320 & AIC & 1.0 & 0.9 & 0.4 & 0.881 & 0.715 & 17.701 & 12.509 \\ 
4 & ESN-0980 & BIC & 1.0 & 0.9 & 0.4 & 0.881 & 0.721 & 17.631 & 12.522 \\ 
5 & ESN-0317 & AIC & 1.0 & 0.8 & 0.4 & 0.883 & 0.713 & 17.599 & 12.600 \\ 
6 & ESN-0321 & AIC & 1.0 & 0.9 & 0.6 & 0.884 & 0.723 & 17.766 & 12.735 \\ 
\rowcolor{light-gray}
7 & ESN-0647 & AICc & 1.0 & 0.8 & 0.4 & 0.884 & \textbf{0.711} & 17.593 & 12.600 \\ 
8 & ESN-0651 & AICc & 1.0 & 0.9 & 0.6 & 0.884 & 0.725 & 17.743 & 12.623 \\ 
9 & ESN-1277 & HQC & 0.9 & 0.9 & 0.4 & 0.884 & 0.713 & 17.594 & 12.571 \\ 
10 & ESN-0318 & AIC & 1.0 & 0.8 & 0.6 & 0.885 & 0.719 & 17.745 & 12.618 \\ 
11 & ESN-0617 & AICc & 0.9 & 0.9 & 0.4 & 0.885 & 0.713 & 17.594 & 12.587 \\ 
12 & ESN-0648 & AICc & 1.0 & 0.8 & 0.6 & 0.885 & 0.718 & 17.723 & 12.532 \\ 
13 & ESN-0947 & BIC & 0.9 & 0.9 & 0.4 & 0.885 & 0.713 & 17.610 & 12.560 \\ 
14 & ESN-1278 & HQC & 0.9 & 0.9 & 0.6 & 0.885 & 0.722 & 17.674 & 12.695 \\ 
15 & ESN-1308 & HQC & 1.0 & 0.8 & 0.6 & 0.885 & 0.716 & 17.685 & 12.556 \\ 
16 & ESN-1311 & HQC & 1.0 & 0.9 & 0.6 & 0.885 & 0.723 & 17.735 & 12.604 \\ 
17 & ESN-1313 & HQC & 1.0 & 1.0 & 0.4 & 0.885 & 0.716 & 17.890 & 12.314 \\ 
18 & ESN-0618 & AICc & 0.9 & 0.9 & 0.6 & 0.886 & 0.721 & 17.671 & 12.763 \\ 
\rowcolor{light-gray}
19 & ESN-0653 & AICc & 1.0 & 1.0 & 0.4 & 0.886 & 0.719 & 17.903 & \textbf{12.254} \\
20 & ESN-0948 & BIC & 0.9 & 0.9 & 0.6 & 0.886 & 0.720 & 17.687 & 12.661 \\ 
21 & ESN-0983 & BIC & 1.0 & 1.0 & 0.4 & 0.886 & 0.720 & 17.898 & 12.337 \\ 
22 & ESN-1307 & HQC & 1.0 & 0.8 & 0.4 & 0.886 & 0.715 & 17.647 & 12.644 \\ 
23 & ESN-0287 & AIC & 0.9 & 0.9 & 0.4 & 0.887 & 0.713 & 17.629 & 12.692 \\ 
24 & ESN-0981 & BIC & 1.0 & 0.9 & 0.6 & 0.887 & 0.721 & 17.732 & 12.623 \\ 
25 & ESN-0285 & AIC & 0.9 & 0.8 & 0.6 & 0.888 & 0.718 & 17.644 & 12.593 \\ 
26 & ESN-0288 & AIC & 0.9 & 0.9 & 0.6 & 0.888 & 0.721 & 17.736 & 12.741 \\ 
27 & ESN-1275 & HQC & 0.9 & 0.8 & 0.6 & 0.889 & 0.717 & 17.672 & 12.509 \\ 
28 & ESN-1280 & HQC & 0.9 & 1.0 & 0.4 & 0.889 & 0.714 & 17.869 & 12.465 \\ 
\rowcolor{light-gray}
29 & ESN-0614 & AICc & 0.9 & 0.8 & 0.4 & 0.890 & 0.720 & \textbf{17.590} & 12.791 \\
30 & ESN-0615 & AICc & 0.9 & 0.8 & 0.6 & 0.890 & 0.718 & 17.673 & 12.584 \\ 
\bottomrule
\end{tabular}
\caption[Top ESN configurations for monthly data]{\textbf{Out-of-sample accuracy} for the 30 best ESN configurations evaluated on \textbf{2,400 monthly time series} in the dataset \textit{Parameter}. For every model, we show the information criterion, leakage rate $\alpha$, spectral radius $\rho$, the reservoir size scaling $\tau$, and the resulting mean/median values of MASE and sMAPE. The rows are sorted by the lowest mean MASE. The minimum value in each accuracy column is highlighted in bold, and the corresponding rows are shaded in light gray.}
\label{tab:pars-monthly-top}
\end{table}

Table \ref{tab:pars-quarterly-top} shows the corresponding results for
the quarterly dataset. Compared to the monthly results, the ranking
structure is more homogeneous: almost all top configurations share
\(\alpha = 1.0\), confirming that when fewer observations per time
series are available, the ESN benefits from updating the internal state
as aggressively as possible rather than retaining long memory. The
optimal spectral radius shifts towards smaller values around
\(\rho = 0.3\) to \(\rho = 0.5\), indicating that more contractive
dynamics are preferable in shorter quarterly time series to prevent
overextension of the internal memory. In contrast to the monthly
dataset, higher \(\tau\)-values (\(\tau = 0.6\)) occur more
systematically among the top-performing configurations. This suggests
that, despite shorter time series, allocating a larger reservoir
compensates for the reduced temporal resolution inherent to quarterly
data. Interestingly, the dispersion across information criteria is again
small, with AIC, AICc, HQC, and BIC all represented in the top ranks,
reinforcing that the criterion for model selection is less critical.
Overall, while both datasets favor high leakage values, the optimal
spectral radius and reservoir scaling clearly depend on data frequency
and series length, supporting the decision to perform a full grid search
rather than fixing hyperparameters a priori.

\begin{table}
\centering
\begin{tabular}{rllrrrrrrr}
\toprule
\multirow{2}{*}{\textbf{Rank}} & \multirow{2}{*}{\textbf{Model}} & \multicolumn{4}{c}{\textbf{Model Specification}} & \multicolumn{2}{c}{\textbf{MASE}} & \multicolumn{2}{c}{\textbf{sMAPE [\%]}} \\
\cmidrule(l){3-6} \cmidrule(l){7-8} \cmidrule(l){9-10}
 & & IC & $\alpha$ & $\rho$ & $\tau$ & Mean & Median & Mean & Median \\
\midrule
\rowcolor{light-gray}
 1 & ESN-0306 & AIC  & 1.0 & 0.4 & 0.6 & \textbf{1.078} & 0.875 & 10.243 & 5.491 \\
 2 & ESN-0636 & AICc & 1.0 & 0.4 & 0.6 & 1.078 & 0.871 & 10.234 & 5.492 \\
 3 & ESN-1296 & HQC  & 1.0 & 0.4 & 0.6 & 1.079 & 0.871 & 10.235 & 5.495 \\
 \rowcolor{light-gray}
 4 & ESN-0305 & AIC  & 1.0 & 0.4 & 0.4 & 1.081 & \textbf{0.861} & 10.240 & 5.430 \\
 5 & ESN-0309 & AIC  & 1.0 & 0.5 & 0.6 & 1.081 & 0.868 & 10.267 & 5.491 \\
 6 & ESN-0966 & BIC  & 1.0 & 0.4 & 0.6 & 1.081 & 0.869 & 10.248 & 5.518 \\
 7 & ESN-1295 & HQC  & 1.0 & 0.4 & 0.4 & 1.081 & 0.863 & 10.228 & 5.441 \\
 8 & ESN-1299 & HQC  & 1.0 & 0.5 & 0.6 & 1.081 & 0.870 & 10.254 & 5.482 \\
 9 & ESN-0303 & AIC  & 1.0 & 0.3 & 0.6 & 1.082 & 0.869 & 10.244 & 5.457 \\
10 & ESN-0633 & AICc & 1.0 & 0.3 & 0.6 & 1.082 & 0.869 & 10.242 & 5.479 \\
11 & ESN-0635 & AICc & 1.0 & 0.4 & 0.4 & 1.082 & 0.861 & 10.244 & 5.450 \\
12 & ESN-0639 & AICc & 1.0 & 0.5 & 0.6 & 1.082 & 0.873 & 10.263 & 5.482 \\
\rowcolor{light-gray}
13 & ESN-1298 & HQC  & 1.0 & 0.5 & 0.4 & 1.082 & 0.872 & \textbf{10.222} & 5.447 \\
14 & ESN-1301 & HQC  & 1.0 & 0.6 & 0.4 & 1.082 & 0.870 & 10.233 & 5.469 \\
15 & ESN-0308 & AIC  & 1.0 & 0.5 & 0.4 & 1.083 & 0.864 & 10.243 & 5.440 \\
16 & ESN-0638 & AICc & 1.0 & 0.5 & 0.4 & 1.083 & 0.864 & 10.242 & 5.440 \\
17 & ESN-0968 & BIC  & 1.0 & 0.5 & 0.4 & 1.083 & 0.867 & 10.231 & 5.463 \\
18 & ESN-0969 & BIC  & 1.0 & 0.5 & 0.6 & 1.083 & 0.873 & 10.267 & 5.535 \\
19 & ESN-0972 & BIC  & 1.0 & 0.6 & 0.6 & 1.083 & 0.875 & 10.271 & 5.503 \\
20 & ESN-0975 & BIC  & 1.0 & 0.7 & 0.6 & 1.083 & 0.871 & 10.252 & 5.522 \\
21 & ESN-1293 & HQC  & 1.0 & 0.3 & 0.6 & 1.083 & 0.869 & 10.248 & 5.454 \\
22 & ESN-1302 & HQC  & 1.0 & 0.6 & 0.6 & 1.083 & 0.869 & 10.273 & 5.514 \\
23 & ESN-0311 & AIC  & 1.0 & 0.6 & 0.4 & 1.084 & 0.863 & 10.251 & 5.495 \\
24 & ESN-0632 & AICc & 1.0 & 0.3 & 0.4 & 1.084 & 0.862 & 10.251 & 5.429 \\
25 & ESN-0642 & AICc & 1.0 & 0.6 & 0.6 & 1.084 & 0.865 & 10.264 & 5.519 \\
26 & ESN-0645 & AICc & 1.0 & 0.7 & 0.6 & 1.084 & 0.869 & 10.272 & 5.528 \\
27 & ESN-0971 & BIC  & 1.0 & 0.6 & 0.4 & 1.084 & 0.875 & 10.240 & 5.490 \\
28 & ESN-0978 & BIC  & 1.0 & 0.8 & 0.6 & 1.084 & 0.867 & 10.273 & 5.480 \\
29 & ESN-1304 & HQC  & 1.0 & 0.7 & 0.4 & 1.084 & 0.865 & 10.265 & 5.612 \\
\rowcolor{light-gray}
30 & ESN-0302 & AIC  & 1.0 & 0.3 & 0.4 & 1.085 & 0.862 & 10.262 & \textbf{5.415} \\
\bottomrule
\end{tabular}
\caption[Top ESN configurations for quarterly data]{\textbf{Out-of-sample accuracy} for the 30 best ESN configurations evaluated on \textbf{1,200 quarterly time series} in the dataset \textit{Parameter}. For every model, we show the information criterion, leakage rate $\alpha$, spectral radius $\rho$, the reservoir size scaling $\tau$, and the resulting mean/median values of MASE and sMAPE. The rows are sorted by the lowest mean MASE. The minimum value in each accuracy column is highlighted in bold, and the corresponding rows are shaded in light gray.}
\label{tab:pars-quarterly-top}
\end{table}

\begin{figure}[hbt!]
\centering
\includegraphics[width=\linewidth]{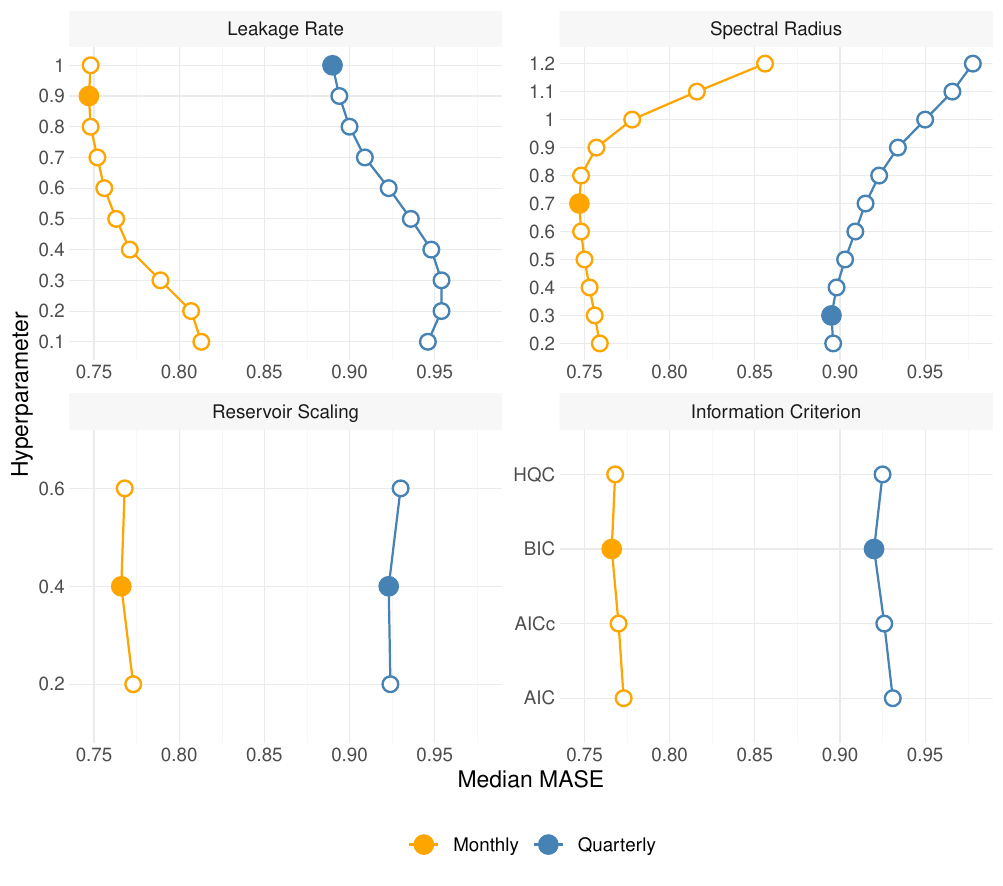}
\caption[Hyperparameter sensitivity summary]{Median MASE across all time series as a function of individual hyperparameter settings, computed by marginalizing over all remaining grid configurations. Monthly results are shown in orange and quarterly results in blue. Each point denotes the median MASE for a specific hyperparameter value; the best-performing value per panel (minimum median MASE) is indicated by a filled marker, while all other values are shown as hollow markers.}
\label{fig:pars-summary}
\end{figure}

To obtain a global view of hyperparameter sensitivity across all series,
Figure \ref{fig:pars-summary} summarizes the median MASE values for each
hyperparameter setting, computed over all models in the grid for both
datasets. The panels are organized by hyperparameter dimension, covering
leakage rate \(\alpha\), spectral radius \(\rho\), reservoir scaling
\(\tau\), and the information criterion. Within each panel, each point
represents the median MASE for a specific hyperparameter value while
marginalizing over all remaining settings, which yields smooth trajectories 
due to aggregation over a large number of time series and model configurations. 
Monthly and quarterly results are shown using a
consistent color scheme (orange for monthly, blue for quarterly). The
optimal value per panel, defined as the minimum median MASE, is
indicated by a filled marker, whereas all remaining values are shown as 
hollow markers. Tables \ref{tab:pars-monthly-dist} and 
\ref{tab:pars-quarterly-dist} in the Appendix report the corresponding
numerical values.

For the monthly dataset, the lowest median MASE is achieved at \(\alpha = 0.9\), \(\rho = 0.7\), \(\tau = 0.4\), and model selection based on BIC. For the quarterly dataset, the best performing values shift toward \(\alpha = 1.0\) and \(\rho = 0.3\), while \(\tau = 0.4\) remains unchanged and BIC again yields the lowest median MASE. These patterns suggest frequency-specific differences in the preferred reservoir dynamics: quarterly time series tend to favor more contractive dynamics and more aggressive updating, whereas monthly data favor slightly more
persistent reservoirs. At the same time, the differences between information criteria are small, suggesting that the choice among AIC, AICc, BIC, and HQC has limited practical impact in this application. Overall, the smoothness of the curves suggests stable empirical tendencies within the considered grid, rather than sharply separated optimal settings.

In summary, the results indicate broadly consistent and interpretable
effects of ESN hyperparameters on forecasting performance, while also
showing that the optimal configurations are not universal across
frequencies. While fast state updates are preferred in both datasets,
differences in spectral radius and reservoir capacity highlight the
importance of adapting ESN dynamics to the temporal resolution of the
data. These findings support the design choice of conducting a 
hyperparameter sweep and provide empirical guidance for selecting ESN
configurations in similar forecasting settings.

\subsection{Random initialization} \label{sec:random-initialization}

Although the ESN hyperparameters determine the overall reservoir dynamics, the model remains stochastic because the reservoir weights are randomly initialized \citep{Lukosevicius2009,
Lukosevicius2012}. Different initializations generate different reservoir-state representations and may therefore lead to different fitted readout weights and recursive forecast paths, even when the data, preprocessing steps, and hyperparameters are held constant. Assessing this source of variation is important because the results of a single ESN fit should not depend excessively on an arbitrary random seed.

To examine the sensitivity of the selected ESN specifications to random initialization, the best-performing monthly and quarterly configurations from the hyperparameter sweep are re-estimated using 30 different random seeds. For the monthly data, all models use AICc, \(\alpha = 1.0\), \(\rho = 0.9\), and \(\tau = 0.4\). For the quarterly data, all models use AIC, \(\alpha = 1.0\), \(\rho = 0.4\), and \(\tau = 0.6\). Thus, the models denoted by ESN-01 to ESN-30 differ only in their random initialization. Seed 42, which was used in the original hyperparameter sweep, is retained as the reference initialization. Each model is evaluated on the complete frequency-specific \textit{Parameter} dataset using MASE and sMAPE.

Tables \ref{tab:rand-monthly-top} and \ref{tab:rand-quarterly-top} summarize the resulting forecast accuracy and rank the initializations by mean MASE. For the monthly data, mean MASE ranges from 0.878 to 0.907, while median MASE ranges from 0.707 to 0.730. The corresponding mean sMAPE values range from 17.592\% to 17.908\%, and median sMAPE ranges from 12.406\% to 13.074\%. Seed 42 achieves the lowest mean MASE and the lowest mean and median sMAPE, whereas seed 229 produces the lowest median MASE. However, the differences between most initializations are small: excluding the lowest-ranked initialization, mean MASE lies between 0.878 and 0.894.

The comparatively high mean MASE of 0.907 for seed 840 is primarily caused by an isolated extreme error rather than a general deterioration across the monthly series. As reported in Appendix Table \ref{tab:rand-monthly-dist}, this initialization produces a maximum MASE of 60.403 and a standard deviation of 1.387. For all other monthly initializations, the maximum MASE ranges from 7.110 to 8.483 and the standard deviation from 0.673 to 0.698. At the same time, seed 840 achieves a mean sMAPE of 17.641\%, which is among the lowest values in Table \ref{tab:rand-monthly-top}. This difference illustrates that aggregate mean accuracy can occasionally be affected by a small number of initialization-specific extreme errors and reinforces the value of considering both mean and median results.

The quarterly results show even less sensitivity to random initialization. Mean MASE varies only between 1.078 and 1.089, while median MASE ranges from 0.860 to 0.884. Mean sMAPE lies between 10.227\% and 10.324\%, and median sMAPE between 5.459\% and 5.578\%. Seeds 42 and 622 achieve the lowest mean MASE after rounding, while seed 602 provides the lowest median MASE. Seed 623 achieves the lowest mean sMAPE, and seed 900 the lowest median sMAPE. Consequently, no single initialization dominates across all accuracy measures. The complete distributions in Appendix Table \ref{tab:rand-quarterly-dist} are also highly similar: the maximum MASE varies only from 7.234 to 7.301 and its standard deviation from 0.824 to 0.838 across the 30 initializations.

Because the selected hyperparameter configurations were originally identified using seed 42 on the same \textit{Parameter} dataset, the strong ranking of this reference seed should not be interpreted as evidence that seed 42 is intrinsically superior. Instead, the relevant finding is the narrow dispersion of forecast accuracy across the alternative seeds. In particular, the quarterly results and the monthly median results indicate that the aggregate performance of the selected ESN configurations is largely stable with respect to random initialization.

\begin{table}[hbt!]
\centering
\begin{tabular}{rlrrrrr}
\toprule
\multirow{2}{*}{\textbf{Rank}} &
\multirow{2}{*}{\textbf{Model}} &
\multirow{2}{*}{\textbf{Seed}} &
\multicolumn{2}{c}{\textbf{MASE}} &
\multicolumn{2}{c}{\textbf{sMAPE [\%]}} \\
\cmidrule(l){4-5} \cmidrule(l){6-7}
& & & Mean & Median & Mean & Median \\
\midrule
1 & ESN-01 & 42 & 0.878 & 0.712 & 17.592 & 12.406 \\ 
2 & ESN-23 & 900 & 0.880 & 0.713 & 17.699 & 12.450 \\ 
3 & ESN-02 & 562 & 0.885 & 0.720 & 17.692 & 12.680 \\ 
4 & ESN-06 & 75 & 0.885 & 0.710 & 17.819 & 12.735 \\ 
5 & ESN-07 & 229 & 0.885 & 0.707 & 17.750 & 12.498 \\ 
6 & ESN-19 & 533 & 0.885 & 0.720 & 17.693 & 12.878 \\ 
7 & ESN-08 & 147 & 0.886 & 0.716 & 17.687 & 12.599 \\ 
8 & ESN-10 & 50 & 0.886 & 0.725 & 17.751 & 12.681 \\ 
9 & ESN-25 & 985 & 0.886 & 0.723 & 17.673 & 12.664 \\ 
10 & ESN-26 & 284 & 0.886 & 0.725 & 17.888 & 12.848 \\ 
11 & ESN-29 & 518 & 0.886 & 0.716 & 17.728 & 12.508 \\ 
12 & ESN-30 & 213 & 0.886 & 0.726 & 17.779 & 12.854 \\ 
13 & ESN-11 & 129 & 0.887 & 0.716 & 17.741 & 12.542 \\ 
14 & ESN-05 & 154 & 0.888 & 0.723 & 17.863 & 12.727 \\ 
15 & ESN-03 & 998 & 0.889 & 0.717 & 17.754 & 12.669 \\ 
16 & ESN-22 & 880 & 0.889 & 0.719 & 17.767 & 12.886 \\ 
17 & ESN-24 & 298 & 0.889 & 0.723 & 17.868 & 12.708 \\ 
18 & ESN-27 & 933 & 0.889 & 0.726 & 17.815 & 12.743 \\ 
19 & ESN-20 & 411 & 0.890 & 0.727 & 17.865 & 12.698 \\ 
20 & ESN-04 & 322 & 0.891 & 0.721 & 17.850 & 12.932 \\ 
21 & ESN-12 & 304 & 0.891 & 0.714 & 17.814 & 12.616 \\ 
22 & ESN-16 & 602 & 0.891 & 0.725 & 17.835 & 12.726 \\ 
23 & ESN-21 & 883 & 0.892 & 0.730 & 17.898 & 13.074 \\ 
24 & ESN-13 & 24 & 0.893 & 0.721 & 17.859 & 12.560 \\ 
25 & ESN-28 & 622 & 0.893 & 0.717 & 17.819 & 12.495 \\ 
26 & ESN-09 & 635 & 0.894 & 0.729 & 17.877 & 12.488 \\ 
27 & ESN-15 & 357 & 0.894 & 0.714 & 17.803 & 12.636 \\ 
28 & ESN-17 & 166 & 0.894 & 0.722 & 17.908 & 12.625 \\ 
29 & ESN-18 & 623 & 0.894 & 0.725 & 17.863 & 12.701 \\ 
30 & ESN-14 & 840 & 0.907 & 0.717 & 17.641 & 12.627 \\ 
\bottomrule
\end{tabular}
\caption[ESN accuracy across random initializations]{\textbf{Out-of-sample accuracy} of the ESN across 30 random initializations evaluated on \textbf{2,400 monthly time series} in the dataset \textit{Parameter}. For each initialization, the table reports the random seed and the resulting mean and median values of MASE and sMAPE. The rows are sorted by mean MASE from low to high.}
\label{tab:rand-monthly-top}
\end{table}

\begin{table}[hbt!]
\centering
\begin{tabular}{rlrrrrr}
\toprule
\multirow{2}{*}{\textbf{Rank}} &
\multirow{2}{*}{\textbf{Model}} &
\multirow{2}{*}{\textbf{Seed}} &
\multicolumn{2}{c}{\textbf{MASE}} &
\multicolumn{2}{c}{\textbf{sMAPE [\%]}} \\
\cmidrule(l){4-5} \cmidrule(l){6-7}
& & & Mean & Median & Mean & Median \\
\midrule
1 & ESN-01 & 42 & 1.078 & 0.875 & 10.243 & 5.491 \\ 
2 & ESN-28 & 622 & 1.078 & 0.871 & 10.241 & 5.488 \\ 
3 & ESN-04 & 322 & 1.079 & 0.866 & 10.253 & 5.480 \\ 
4 & ESN-09 & 635 & 1.079 & 0.867 & 10.253 & 5.500 \\ 
5 & ESN-19 & 533 & 1.079 & 0.863 & 10.252 & 5.505 \\ 
6 & ESN-02 & 562 & 1.080 & 0.869 & 10.249 & 5.533 \\ 
7 & ESN-27 & 933 & 1.080 & 0.867 & 10.249 & 5.479 \\ 
8 & ESN-11 & 129 & 1.081 & 0.864 & 10.254 & 5.476 \\ 
9 & ESN-05 & 154 & 1.082 & 0.866 & 10.252 & 5.578 \\ 
10 & ESN-16 & 602 & 1.082 & 0.860 & 10.281 & 5.463 \\ 
11 & ESN-18 & 623 & 1.082 & 0.864 & 10.227 & 5.462 \\ 
12 & ESN-25 & 985 & 1.082 & 0.862 & 10.253 & 5.472 \\ 
13 & ESN-08 & 147 & 1.083 & 0.875 & 10.277 & 5.493 \\ 
14 & ESN-14 & 840 & 1.083 & 0.874 & 10.264 & 5.560 \\ 
15 & ESN-20 & 411 & 1.083 & 0.877 & 10.274 & 5.505 \\ 
16 & ESN-12 & 304 & 1.084 & 0.869 & 10.262 & 5.478 \\ 
17 & ESN-13 & 24 & 1.084 & 0.868 & 10.261 & 5.517 \\ 
18 & ESN-15 & 357 & 1.084 & 0.867 & 10.264 & 5.515 \\ 
19 & ESN-07 & 229 & 1.085 & 0.872 & 10.263 & 5.530 \\ 
20 & ESN-10 & 50 & 1.085 & 0.866 & 10.311 & 5.508 \\ 
21 & ESN-17 & 166 & 1.085 & 0.866 & 10.299 & 5.510 \\ 
22 & ESN-06 & 75 & 1.086 & 0.866 & 10.294 & 5.483 \\ 
23 & ESN-21 & 883 & 1.086 & 0.864 & 10.287 & 5.478 \\ 
24 & ESN-22 & 880 & 1.086 & 0.871 & 10.308 & 5.484 \\ 
25 & ESN-23 & 900 & 1.086 & 0.870 & 10.303 & 5.459 \\ 
26 & ESN-24 & 298 & 1.086 & 0.871 & 10.324 & 5.556 \\ 
27 & ESN-30 & 213 & 1.086 & 0.868 & 10.282 & 5.489 \\ 
28 & ESN-03 & 998 & 1.088 & 0.871 & 10.284 & 5.555 \\ 
29 & ESN-26 & 284 & 1.088 & 0.884 & 10.311 & 5.523 \\ 
30 & ESN-29 & 518 & 1.089 & 0.871 & 10.273 & 5.503 \\ 
\bottomrule
\end{tabular}
\caption[ESN accuracy across random initializations]{\textbf{Out-of-sample accuracy} of the ESN across 30 random initializations evaluated on \textbf{1,200 quarterly time series} in the dataset \textit{Parameter}. For each initialization, the table reports the random seed and the resulting mean and median values of MASE and sMAPE. The rows are sorted by mean MASE from low to high.}
\label{tab:rand-quarterly-top}
\end{table}

\begin{figure}[hbt!]
\centering
\includegraphics[width=\linewidth]{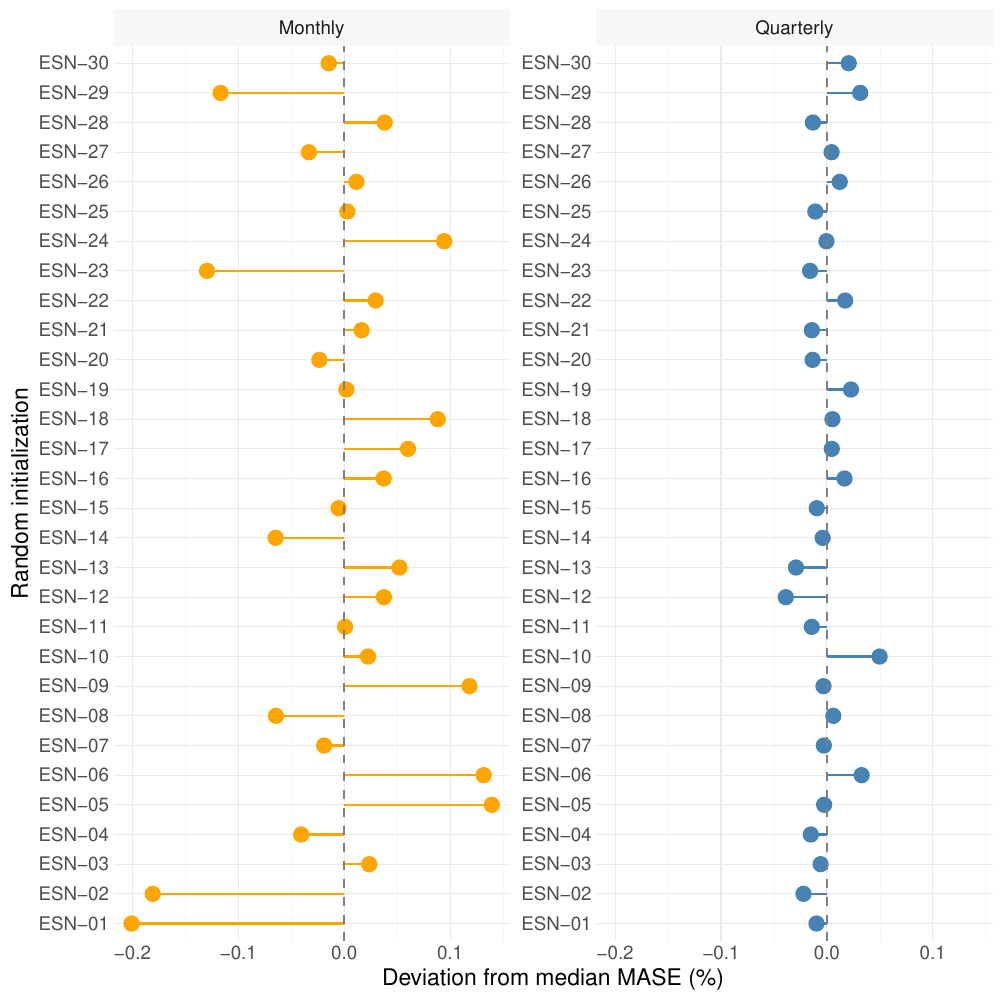}
\caption[Sensitivity to random initialization]{Median deviation from the series-specific median MASE across random initializations. For each time series, the MASE obtained from a given initialization is expressed as a percentage deviation from the median MASE across all tested initializations. Monthly results are shown in orange and quarterly results in blue. Each point represents the median deviation across all time series for a specific random initialization. Negative values indicate better-than-median forecast accuracy, whereas positive values indicate worse-than-median forecast accuracy.}
\label{fig:rand-summary}
\end{figure}

Figure \ref{fig:rand-summary} provides a complementary series-level comparison. For each time series, the MASE obtained from an initialization is expressed as a percentage deviation from the median MASE across all tested initializations for that series. This normalization controls for differences in forecasting difficulty between time series. The seed-specific median deviations remain concentrated around the zero-reference line for both frequencies, indicating that no initialization produces a consistently large improvement or deterioration across the complete dataset. Negative values identify seeds that perform slightly better than the series-specific median on average, whereas positive values indicate slightly worse performance. The limited spread around zero is consistent with the narrow ranges reported in Tables \ref{tab:rand-monthly-top} and \ref{tab:rand-quarterly-top}.

\begin{figure}[hbt!]
\centering
\includegraphics[width=\linewidth]{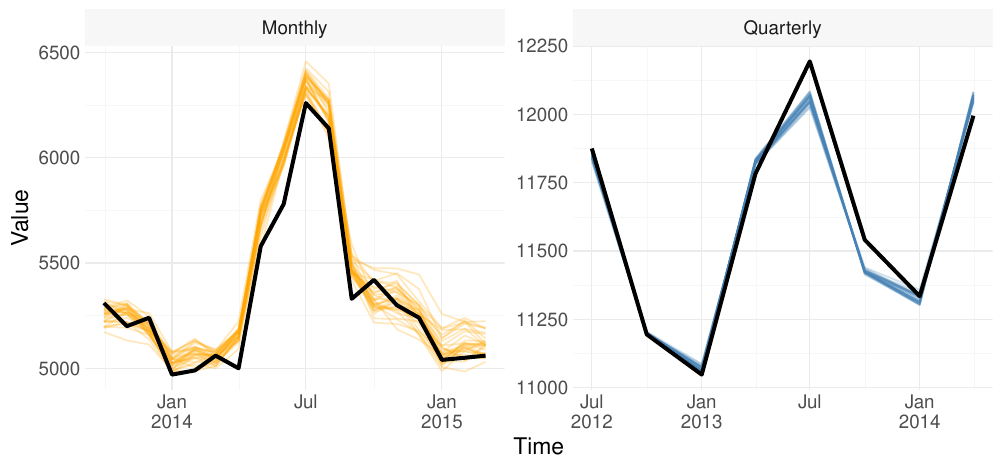}
\caption[Forecasts under varying random initializations]{Forecasts under varying random initializations of the ESN. Left: forecasts for the example monthly series M21655 over an 18-period forecast horizon. Right: forecasts for the example quarterly series Q13921 over an 8-period forecast horizon. All frequency-specific hyperparameters are held constant, while only the random seed used to initialize the reservoir is varied. Actual observations (holdout data) are shown in black. Monthly forecasts are shown in orange and quarterly forecasts in blue, illustrating the sensitivity of the resulting forecasts to random initialization.}
\label{fig:fcst-rand}
\end{figure}

While aggregate accuracy is comparatively stable, individual forecast paths can still differ across reservoir realizations. Figure \ref{fig:fcst-rand} illustrates this effect for the monthly series M21655 and the quarterly series Q13921. All preprocessing choices and frequency-specific hyperparameters are fixed, and only the random initialization is varied. The resulting forecast paths are therefore not identical and may differ in their level, responsiveness, and development over the recursive forecast horizon. This variation arises because each reservoir generates a different nonlinear representation of the same input history. The figure demonstrates that random initialization can matter for an individual time series even when its effect largely averages out across a large collection of series.

Overall, the analysis indicates that the selected ESN specifications are reasonably robust to random initialization at the aggregate level. The exact accuracy values and forecasts for individual series may change with the random seed, and isolated unfavorable reservoir realizations can produce large errors. Nevertheless, the small differences in mean and median accuracy across most seeds suggest that the main empirical conclusions are not driven by a broadly unstable initialization. Fixing the random seed remains necessary for reproducibility, while averaging forecasts across several independently initialized reservoirs could provide an additional way to reduce the remaining initialization-specific variation (i.e., ensemble forecasting).

\subsection{Benchmark analysis} \label{sec:benchmark-analysis}

In order to evaluate the performance of ESNs, forecast accuracy and computational runtime are compared against simple, statistical, and NN forecasting methods. The choice of benchmark methods is mainly driven by their relevance to automated forecasting and their use in forecasting research and practice. In addition to established simple and statistical approaches, the comparison includes the MLP and RNN benchmarks from the M4 Forecasting Competition and the competition winning model ES-RNN.

The reported ESN runtime refers to training and forecasting on the
\textit{Forecast} dataset after the main ESN hyperparameters
\(\alpha\), \(\rho\), and \(\tau\), as well as the information criterion,
have been selected on the separate \textit{Parameter} dataset. The ridge
penalty \(\lambda\), however, is still selected within each fitted ESN
by random search using the chosen information criterion. The reported
runtime therefore includes ESN training, the internal selection of
\(\lambda\), and recursive forecasting for the selected ESN
specification, but not the preceding large-scale hyperparameter sweep on
the \textit{Parameter} dataset. Importantly, the MLP, RNN, and ES-RNN
forecasts were taken from the published M4 benchmark and competition
results and were not re-trained in the computational environment used
for this study. Consequently, the runtime comparison is restricted to
the ESN and the locally estimated simple and statistical benchmark
methods and should not be interpreted as a complete computational
comparison across all evaluated forecasting methods.

Table \ref{tab:benchmark-methods} summarizes the simple, statistical, and NN methods used in the forecast evaluation. The simple methods provide transparent reference points with different assumptions about the underlying time series. The seasonal naive method, for example, reproduces the most recent observation from the corresponding season and can perform well when seasonality is strong and other systematic changes are limited. Naive2 extends the naive approach by first removing seasonality when it is statistically detected and subsequently restoring the estimated seasonal pattern to the forecasts. The drift method allows the most recent observation to increase or decrease according to the average historical change and may therefore be more appropriate for trending series \citep{Hyndman2021, Makridakis1998, Makridakis2020}.

The statistical benchmarks include automatic ARIMA and ETS models, which are widely used in empirical forecasting studies \citep{Hyndman2008a, Hyndman2008b, Makridakis2020}, as well as the Theta method \citep{Assimakopoulos2000} and TBATS \citep{DeLivera2011}. In addition, the analysis includes the MLP and RNN benchmarks used in the M4 Forecasting Competition. Both NN benchmarks are fitted separately to each time series using lagged observations after detrending and seasonal adjustment, with forecasts generated recursively over the required horizon \citep{Makridakis2020}. Finally, ES-RNN is included as a more advanced hybrid benchmark. This method combines series-specific exponential smoothing components with an long short-term memory (LSTM) network trained across multiple time series and constituted the winning method of the M4 Forecasting Competition \citep{Smyl2020}.

The simple and statistical benchmark methods are estimated locally,
whereas the MLP, RNN, and ES-RNN forecasts are taken from the
corresponding published M4 benchmark and competition results
\citep{Makridakis2020, Smyl2020} and were not re-trained locally.
Consequently, comparable computational runtimes are not available for
these three methods. Details on the software packages, computational
workflow, and computing environment are provided in Appendix
\hyperref[sec:appendix-computational]{A.5}.

\begin{table}[hbt!]
\centering
\begin{tabular}{p{1.5cm} p{14.0cm}}
\toprule
\textbf{Model} & \textbf{Description} \\
\midrule
NAIVE & Naive forecast (random walk model). All forecasts are set to the value of the last observation of the training data. \\
\midrule
NAIVE2 & Naive forecast applied to seasonally adjusted data when seasonality is detected. The forecasts are subsequently transformed back by restoring the estimated seasonal pattern. \\
\midrule
DRIFT & Naive forecast plus drift (random walk with drift). A variation of the naive method, where the forecast is allowed to increase or decrease over time. The amount of change over time (drift) is set to be the average change in the training data. \\
\midrule
SNAIVE & Seasonal naive forecast. Each forecast is set to be equal to the last observed value from the same season (e.g., for monthly data the same month of the previous year). \\
\midrule
MEAN & Mean forecast. All forecasts are set to the arithmetic mean (average) of the training data. \\
\midrule
ARIMA & Automatic ARIMA model using the Hyndman-Khandakar algorithm. Searches through the model space ARIMA$(p,d,q)(P,D,Q)_m$ to identify the best seasonal or nonseasonal ARIMA model with the lowest AICc value. \\
\midrule
ETS & Automatic exponential smoothing state space model, based on the classification of exponential smoothing methods (Error, Trend, Season). For example, ETS$(M,A,A)$ is Holt-Winters' method with multiplicative error, additive trend and additive seasonality. Model estimation is done via log-likelihood and model selection via AICc. \\
\midrule
THETA & Theta method of Assimakopoulos and Nikolopoulos (2000). Theta is equivalent to simple exponential smoothing (ETS$(A,N,N)$) with drift applied to a nonseasonal or seasonally adjusted series. \\
\midrule
TBATS & Exponential smoothing state space model with trigonometric seasonality, Box-Cox transformation, ARMA errors, trend and seasonal components. Originally intended for high-frequency time series with multiple and nested seasonal patterns. \\
\midrule
MLP & Multilayer perceptron benchmark with a single feed-forward hidden layer. The model uses lagged, detrended, and seasonally adjusted observations as inputs and generates forecasts recursively. \\
\midrule
RNN & Simple recurrent neural network benchmark with a recurrent hidden layer and linear output. The model uses lagged, detrended, and seasonally adjusted observations and generates forecasts recursively. \\
\midrule
ES-RNN & Hybrid exponential smoothing and recurrent neural network model. Exponential smoothing represents series-specific level and seasonality, while an LSTM network learns common nonlinear patterns across time series. \\
\bottomrule
\end{tabular}
\caption[Benchmark methods]{\textbf{Benchmark methods:} Overview and description of the simple, statistical, and neural network methods used for the comparison of forecast accuracy.}
\label{tab:benchmark-methods}
\end{table}

In the benchmark analysis, the forecast accuracy of the benchmark
methods listed in Table \ref{tab:benchmark-methods} is compared with the
best-performing ESN configurations identified in the hyperparameter
sweep. Specifically, these are ESN-0650 for the monthly data (AICc,
\(\alpha = 1.0\), \(\rho = 0.9\), \(\tau = 0.4\)) and ESN-0306 (AIC,
\(\alpha = 1.0\), \(\rho = 0.4\), \(\tau = 0.6\)) for the quarterly
data.

Tables \ref{tab:metrics-monthly} and \ref{tab:metrics-quarterly}
summarize the results for the monthly and quarterly datasets regarding
forecast accuracy and computational runtime. Forecast accuracy is
reported using OWA and the arithmetic mean and median of MASE and sMAPE
across all time series of the respective frequency. Computational
runtime is reported as total runtime for the entire monthly or quarterly
dataset and average runtime per time series, measured in seconds. 
Runtime values are not available for MLP, RNN, and ES-RNN because their forecasts are taken
from the published M4 benchmark and competition results. The models are
ranked by OWA, with lower values indicating better overall forecast
accuracy. The best value in each column is highlighted in bold, and the
ESN is marked in light gray.

Table \ref{tab:metrics-monthly} reports the forecasting performance of
13 models on 2,400 monthly univariate time series. ES-RNN achieves the
lowest OWA (0.858) and the best value for all four reported accuracy
statistics, with a mean MASE of 0.870, a median MASE of 0.679, a mean
sMAPE of 15.746\%, and a median sMAPE of 11.579\%. Among the statistical
models, TBATS attains the lowest OWA (0.898), followed by Theta (0.902),
ARIMA (0.914), and ETS (0.916). ESN-0650 ranks sixth overall with an OWA
of 0.918. Nevertheless, its mean MASE of 0.898 is nearly identical to
those of ARIMA (0.897) and TBATS (0.899), while its mean sMAPE of
17.463\% is higher than the values obtained by the leading methods.
Regarding computational efficiency, Naive2 and the other simple methods
have the lowest runtimes, whereas TBATS is by far the slowest method.
ESN-0650 requires approximately 0.36 seconds per time series and is
faster than ARIMA and TBATS, but slower than ETS and the simple
benchmarks.

The monthly results indicate that the ESN remains competitive with the
established statistical models, particularly under MASE. Although it
does not lead the OWA ranking in Table \ref{tab:metrics-monthly}, the
differences between ESN-0650, ARIMA, and TBATS in mean MASE are very
small. Its lower OWA rank is mainly attributable to its comparatively
weaker sMAPE performance, since OWA combines the relative mean MASE and
mean sMAPE results. ES-RNN, however, outperforms the ESN and all other
evaluated methods across the reported aggregate accuracy measures. The
computational profile of the ESN remains favorable relative to ARIMA and
especially TBATS. Overall, the monthly results support the ESN as a
competitive and computationally efficient alternative to established
statistical methods, but not as the best-performing method once ES-RNN
and the combined OWA measure are considered.

\begin{table}[hbt!]
\centering
\begin{tabular}{lrrrrrrr}
\toprule
\multirow{2}{*}{\textbf{Model}} &
\multirow{2}{*}{\textbf{OWA}} &
\multicolumn{2}{c}{\textbf{MASE}} &
\multicolumn{2}{c}{\textbf{sMAPE [\%]}} &
\multicolumn{2}{c}{\textbf{Runtime [sec]}} \\
\cmidrule(l){3-4} \cmidrule(l){5-6} \cmidrule(l){7-8}
& & Mean & Median & Mean & Median & Mean & Total \\
\midrule
ES-RNN   & \textbf{0.858} & \textbf{0.870} & \textbf{0.679} &
\textbf{15.746} & \textbf{11.579} & -- & -- \\
TBATS    & 0.898 & 0.899 & 0.718 & 16.686 & 11.845 & 0.997 & 2393.176 \\
THETA    & 0.902 & 0.911 & 0.724 & 16.638 & 11.666 & 0.036 & 86.347 \\
ARIMA    & 0.914 & 0.897 & 0.733 & 17.334 & 11.845 & 0.411 & 986.246 \\
ETS      & 0.916 & 0.905 & 0.713 & 17.257 & 11.696 & 0.279 & 668.774 \\
\rowcolor{light-gray}
ESN-0650 & 0.918 & 0.898 & 0.723 & 17.463 & 12.109 & 0.364 & 872.552 \\
NAIVE2   & 1.000 & 0.999 & 0.803 & 18.635 & 13.097 & \textbf{0.027} & \textbf{64.965} \\
NAIVE    & 1.076 & 1.110 & 0.859 & 19.386 & 14.387 & 0.029 & 70.492 \\
DRIFT    & 1.080 & 1.101 & 0.835 & 19.716 & 13.263 & 0.029 & 70.788 \\
SNAIVE   & 1.152 & 1.195 & 0.967 & 20.642 & 15.602 & 0.029 & 70.682 \\
MLP      & 1.599 & 1.544 & 1.147 & 30.791 & 18.725 & -- & -- \\
RNN      & 1.617 & 1.514 & 1.145 & 32.041 & 18.959 & -- & -- \\
MEAN     & 2.462 & 2.893 & 1.992 & 37.797 & 33.115 & 0.028 & 68.006 \\
\bottomrule
\end{tabular}
\caption[Monthly benchmark accuracy and runtime]{\textbf{Monthly dataset:}
OWA, mean and median MASE and sMAPE, and computational runtime for the
forecasting models evaluated on the 2,400 monthly time series in the
dataset \textit{Forecast}. The models are ranked according to OWA from
low to high. The best-performing method (minimum value) per column is
highlighted in bold, and the ESN is shaded in light gray. Runtime values
are unavailable for ES-RNN, MLP, and RNN.}
\label{tab:metrics-monthly}
\end{table}

Table \ref{tab:metrics-quarterly} presents forecasting accuracy results
for 13 models evaluated on 1,200 quarterly time series. ES-RNN achieves
the lowest OWA (0.883) and the best value for all four reported accuracy
statistics, with a mean MASE of 1.097, a median MASE of 0.813, a mean
sMAPE of 10.502\%, and a median sMAPE of 5.344\%. ETS ranks second
according to OWA (0.900), followed by ESN-0306 (0.910), ARIMA (0.917),
and TBATS (0.921). ESN-0306 achieves the second-lowest mean MASE
(1.111) and the second-lowest median sMAPE (5.701\%), while ETS provides
a slightly lower mean sMAPE and median MASE. Regarding computational
efficiency, the simple benchmarks are the fastest methods for which
runtimes are available. TBATS is the slowest method, while ESN-0306
requires approximately 0.14 seconds per time series and has a runtime
similar to ARIMA.

The quarterly results again demonstrate that the ESN yields competitive
accuracy relative to the statistical benchmarks. It ranks third
according to OWA in Table \ref{tab:metrics-quarterly} and performs
particularly well under mean MASE and median sMAPE. However, ES-RNN
performs better across all reported aggregate accuracy measures, while
ETS achieves a slightly lower OWA than the ESN. The results therefore 
show that the ESN is among the strongest methods considered for the 
quarterly data, although ES-RNN and ETS achieve better aggregate 
performance according to OWA. In terms of
computational efficiency, the ESN is substantially faster than TBATS,
although its runtime is slightly higher than those of ARIMA and ETS.
Overall, the quarterly results indicate a favorable balance between
forecast accuracy and computational cost, while also demonstrating the
advantage of the more advanced globally trained ES-RNN model.

\begin{table}[hbt!]
\centering
\begin{tabular}{lrrrrrrr}
\toprule
\multirow{2}{*}{\textbf{Model}} &
\multirow{2}{*}{\textbf{OWA}} &
\multicolumn{2}{c}{\textbf{MASE}} &
\multicolumn{2}{c}{\textbf{sMAPE [\%]}} &
\multicolumn{2}{c}{\textbf{Runtime [sec]}} \\
\cmidrule(l){3-4} \cmidrule(l){5-6} \cmidrule(l){7-8}
& & Mean & Median & Mean & Median & Mean & Total \\
\midrule
ES-RNN   & \textbf{0.883} & \textbf{1.097} & \textbf{0.813} &
\textbf{10.502} & \textbf{5.344} & -- & -- \\
ETS      & 0.900 & 1.121 & 0.868 & 10.689 & 5.805 & 0.076 & 91.462 \\
\rowcolor{light-gray}
ESN-0306 & 0.910 & 1.111 & 0.875 & 11.023 & 5.701 & 0.136 & 163.243 \\
ARIMA    & 0.917 & 1.139 & 0.872 & 10.917 & 5.875 & 0.123 & 147.634 \\
TBATS    & 0.921 & 1.160 & 0.879 & 10.832 & 6.028 & 0.505 & 605.496 \\
THETA    & 0.930 & 1.171 & 0.922 & 10.932 & 6.186 & 0.032 & 38.478 \\
DRIFT    & 0.972 & 1.192 & 0.908 & 11.706 & 6.005 & 0.028 & 33.692 \\
NAIVE2   & 1.000 & 1.286 & 1.027 & 11.519 & 6.697 & 0.027 & 32.683 \\
NAIVE    & 1.040 & 1.347 & 1.082 & 11.888 & 7.145 & 0.032 & 38.175 \\
SNAIVE   & 1.164 & 1.522 & 1.281 & 13.185 & 8.194 & 0.028 & 34.031 \\
MLP      & 1.512 & 1.935 & 1.454 & 17.505 & 9.866 & -- & -- \\
RNN      & 1.539 & 1.995 & 1.485 & 17.584 & 10.048 & -- & -- \\
MEAN     & 2.974 & 4.240 & 3.521 & 30.522 & 25.888 &
\textbf{0.027} & \textbf{32.320} \\
\bottomrule
\end{tabular}
\caption[Quarterly benchmark accuracy and runtime]{\textbf{Quarterly dataset:}
OWA, mean and median MASE and sMAPE, and computational runtime for the
forecasting models evaluated on the 1,200 quarterly time series in the
dataset \textit{Forecast}. The models are ranked according to OWA from
low to high. The best-performing method (minimum value) per column is
highlighted in bold, and the ESN is shaded in light gray. Runtime values
are unavailable for ES-RNN, MLP, and RNN.}
\label{tab:metrics-quarterly}
\end{table}

\begin{sidewaysfigure}
\centering

\begin{subfigure}[t]{0.49\textheight}
  \centering
  \includegraphics[width=\linewidth]
    {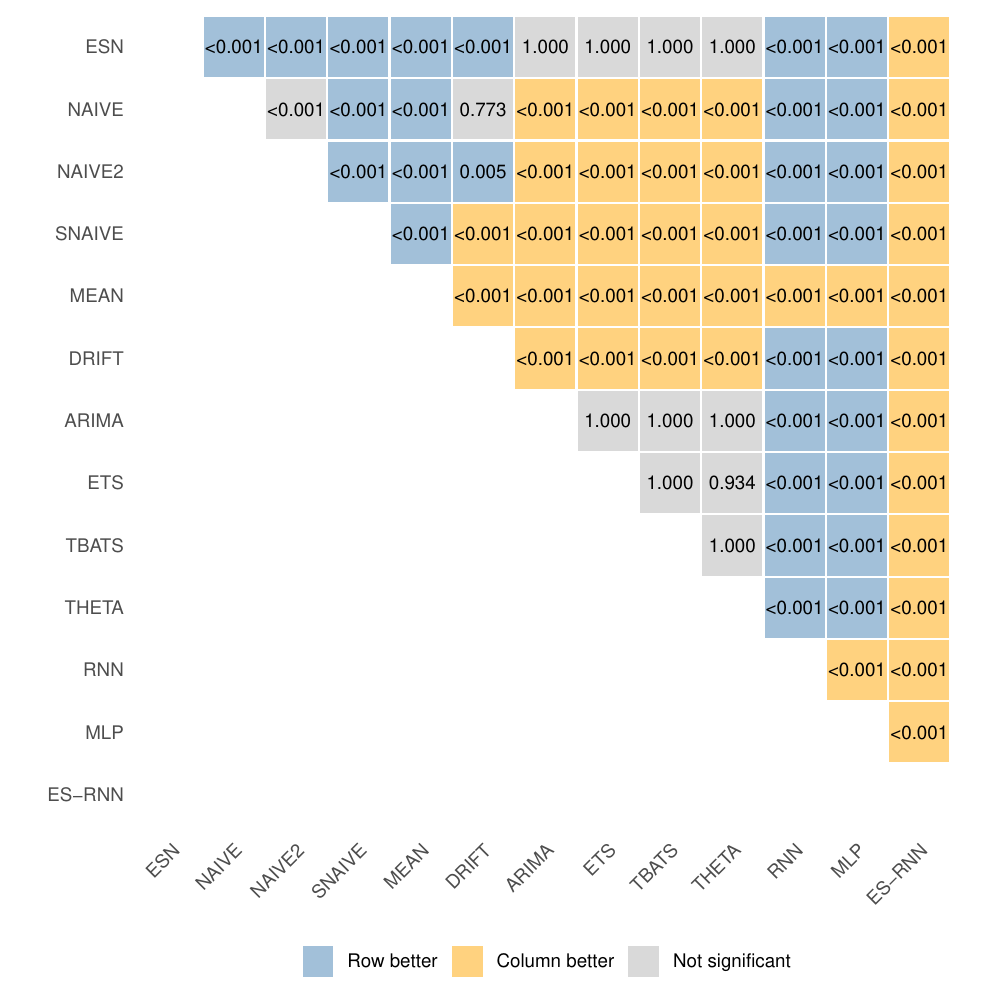}
  \caption{Monthly dataset}
\end{subfigure}
\hfill
\begin{subfigure}[t]{0.49\textheight}
  \centering
  \includegraphics[width=\linewidth]
    {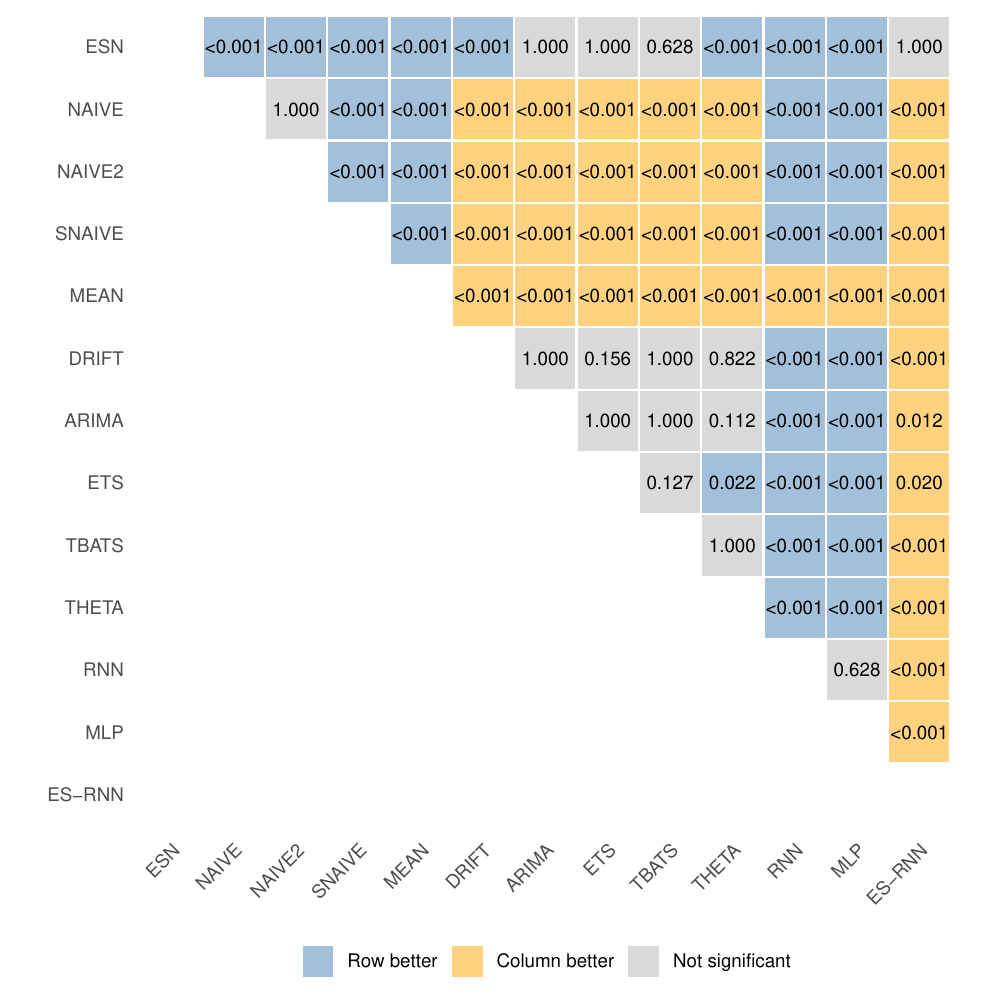}
  \caption{Quarterly dataset}
\end{subfigure}

\caption[Pairwise Wilcoxon tests for forecast accuracy]{Pairwise comparisons of forecasting models based on MASE. The cells report Holm-adjusted p-values from paired Wilcoxon signed-rank tests across the individual time series. Blue cells indicate that the row model has significantly lower MASE than the column model, orange cells indicate that the column model has significantly lower MASE, and gray cells indicate that no statistically significant difference is detected.}
\label{fig:wilcoxon-mase}

\end{sidewaysfigure}

Figure \ref{fig:wilcoxon-mase} complements the descriptive accuracy results with pairwise statistical comparisons based on the MASE values for the individual time series. For each pair of forecasting methods, a paired Wilcoxon signed-rank test \citep{Wilcoxon1945} is used to examine whether the distribution of the series-level MASE differences is centered around zero. Because multiple pairwise comparisons are conducted, the resulting \(p\)-values are adjusted using the Holm procedure \citep{Holm1979}. The tests therefore assess whether the observed differences between methods are systematic across the time series rather than being driven only by differences in aggregate mean accuracy.

For the monthly data, the ESN achieves significantly lower MASE than the naive method, Naive2, the seasonal naive method, the mean and drift forecasts, as well as the MLP and RNN benchmarks. By contrast, no statistically significant difference is detected between the ESN and ARIMA, ETS, TBATS, or the Theta method. These results support the conclusion drawn from the descriptive statistics that the ESN performs at a level comparable to the strongest statistical benchmarks for monthly forecasting. The ES-RNN is the only model that achieves significantly lower monthly MASE than the ESN. However, the corresponding median difference is comparatively small, indicating that statistical significance does not necessarily imply a large practical difference in forecast accuracy.

For the quarterly data, the ESN significantly outperforms the naive, Naive2, 
seasonal naive, mean, drift, Theta, MLP, and RNN methods. No
statistically significant differences are found relative to ARIMA, ETS,
TBATS, or ES-RNN. Thus, although the ESN achieves the second-lowest mean
MASE in Table \ref{tab:metrics-quarterly}, the pairwise tests do not
establish that it is systematically more accurate than these closely
competing models across the individual quarterly series. Taken together
with Figure \ref{fig:wilcoxon-mase}, the results indicate that the ESN
clearly improves upon the simpler benchmarks and the standalone
neural-network models, while its performance is generally statistically
indistinguishable from that of the strongest statistical methods and
ES-RNN. The exact median differences and Holm-adjusted \(p\)-values for
comparisons involving the ESN are reported in Appendix Tables
\ref{tab:wilcoxon-monthly} and \ref{tab:wilcoxon-quarterly}.

Overall, the results in Tables \ref{tab:metrics-monthly} and
\ref{tab:metrics-quarterly} show that the proposed autoregressive ESN is
empirically competitive with established statistical forecasting
methods in the considered setting. Its performance is particularly
strong under MASE, whereas its ranking is less favorable under sMAPE
and the combined OWA measure. ES-RNN achieves the best aggregate
accuracy for both frequencies, while Figure \ref{fig:wilcoxon-mase}
shows that the ESN is generally statistically indistinguishable from the
strongest statistical benchmarks. Thus, the results support the ESN as
a competitive and computationally efficient alternative to standard
statistical methods, but not as a uniformly superior approach across all
models, metrics, and frequencies. Additional detail is provided in
Appendix Tables \ref{tab:mase-monthly}, \ref{tab:smape-monthly},
\ref{tab:mase-quarterly}, and \ref{tab:smape-quarterly}, which report
the minimum, first quartile, mean, median, third quartile, maximum, and
standard deviation of the forecast errors.

Before summarizing the main findings, several aspects of the empirical
design should be kept in mind. The analysis is based on filtered monthly
and quarterly M4 series with at most 20 years of history, so the results
refer to this forecasting setting rather than to the full M4 benchmark.
Moreover, the ESN specification is intentionally simple and uses only
the first autoregressive lag as input; richer input structures,
including explicit seasonal lags, may lead to different results. The
comparison includes the MLP, RNN, and ES-RNN models from the M4
Forecasting Competition, but it does not provide a systematic evaluation
of newer neural forecasting architectures or more advanced reservoir
computing variants. Finally, although pairwise statistical comparisons
are reported for MASE, the results remain conditional on the reservoir
initialization scheme and the set of forecasting methods considered in
this study.

\section{Summary and concluding remarks} \label{sec:summary}

This paper investigated the use of ESNs for univariate time series
forecasting in a large-scale setting. Building on the M4 Forecasting
Competition dataset \citep{Makridakis2020}, we focused on monthly and
quarterly series with at most 20 years of historical data to reflect
realistic constraints found in many organizational environments, where
long histories are often unavailable or only partially relevant. To
ensure that the empirical analysis remained both rigorous and
practically meaningful, the dataset was split into two randomly drawn
subsets: a \textit{Parameter} dataset for systematic ESN hyperparameter
exploration and a disjoint \textit{Forecast} dataset for out-of-sample
evaluation against benchmark methods. An exploratory data analysis
showed that both subsets preserve several important characteristics of
the filtered monthly and quarterly M4 series, while also reflecting the
imposed restriction on the time series length.

Methodologically, the study proposed a first-order autoregressive ESN
framework tailored to univariate forecasting. The approach combines
standard preprocessing steps like stationarity testing via the KPSS
test, differencing when required, and scaling to a symmetric interval
with a leaky integrator ESN whose reservoir size is scaled with the time
series length and capped to avoid excessive complexity. The internal
reservoir dynamics are governed by three key hyperparameters: leakage
rate, spectral radius, and reservoir scaling. The linear readout is
trained by ridge regression, where the regularization parameter is
selected via random search guided by information criteria. This design
provides a flexible yet conceptually simple forecasting pipeline that
can be applied in a fully automated fashion across large collections of
time series.

A central contribution of the paper is the comprehensive hyperparameter
sweep, which systematically evaluated the impact of leakage rate,
spectral radius, reservoir size, and information criterion across
thousands of time series and more than four million ESN fits. The
results show that ESN hyperparameters exert broadly consistent and
interpretable effects on forecasting performance. For monthly data, high
leakage rates (\(\alpha \approx 0.9-1.0\)), spectral radii close to the
edge of stability (\(\rho \approx 0.8-1.0\)), and moderate reservoir
scaling (\(\tau \approx 0.4\)) yielded the best overall performance. For
quarterly data, optimal configurations shifted towards even higher
leakage (\(\alpha = 1.0\)) and more contractive dynamics (spectral
radius around \(\rho = 0.3-0.5\)), while reservoir scaling tended to be
somewhat larger. These patterns reflect differences in temporal
resolution and effective memory requirements between monthly and
quarterly time series. The smoothness of the hyperparameter response
curves suggests that the identified settings reflect stable empirical
tendencies within the considered grid rather than sharply separated
optimal configurations.

The empirical benchmark analysis on the \textit{Forecast} dataset
further demonstrated that the proposed ESN can compete with
well-established statistical models in terms of forecast accuracy while
offering favorable computational properties. For the monthly series, the
best ESN configuration achieved a mean MASE that was nearly identical
to those of ARIMA and TBATS, although ES-RNN provided the best aggregate
accuracy according to OWA, MASE, and sMAPE. For the quarterly series,
the ESN achieved the second-lowest mean MASE and the second-lowest median
sMAPE and ranked third according to OWA, behind ES-RNN and ETS. ES-RNN
again provided the best aggregate accuracy, whereas the standalone MLP
and RNN benchmarks performed less favorably. Pairwise Wilcoxon
signed-rank tests showed that the ESN clearly outperformed the simpler
benchmarks and the standalone NN models, while its MASE was generally
statistically indistinguishable from that of the strongest statistical
methods. Across both frequencies, the selected ESN specification could be trained
and applied efficiently on the \textit{Forecast} dataset. These reported
runtimes represent deployment after the main ESN configuration has been
selected and exclude the preceding hyperparameter sweep comprising
4,752,000 ESN fits on the \textit{Parameter} dataset, for which a
reliable total runtime was not recorded. In addition, the MLP, RNN, and
ES-RNN models were not re-trained locally, and comparable runtimes are
therefore not available for these methods. Consequently, the reported
runtime results should not be interpreted as evidence that the complete
ESN model-selection pipeline is computationally more efficient than all
evaluated alternatives. Taken together, the results indicate that simple
ESNs can offer a useful trade-off between forecast accuracy and
deployment cost once the main hyperparameters have been selected.

At the same time, several limitations of the present study highlight
opportunities for future research. First, the analysis is restricted to
univariate, autoregressive ESNs without exogenous inputs. Many
real-world forecasting problems involve explanatory variables such as
prices, promotions, or macroeconomic indicators; extending the ESN
framework to multivariate or input-driven architectures would therefore
be a natural next step. Second, we focused on point forecasts only and
evaluated performance using sMAPE, MASE, and OWA. In many applications,
decision makers require full predictive distributions or interval
forecasts; integrating probabilistic output layers or ensemble
techniques into the ESN framework could address this need
\citep{Xu2016, Yao2019}. Third, the study considered a single class of
reservoir architectures with \(\tanh(\cdot)\) activation, fixed
sparsity, and basic random initialization. Alternative reservoir
designs, such as modular reservoirs \citep{Babinec2012}, as well as deep
learning approaches (i.e., multiple reservoirs)
\citep{Gallicchio2017, Gallicchio2018}, may further enhance performance
or stability and merit systematic investigation.

Moreover, while the present work employed an exhaustive grid search for
core ESN hyperparameters and random search for the ridge penalty, more
advanced optimization strategies could be explored. Bayesian
optimization \citep{Maat2018} or Particle Swarm Optimization
\citep{Chouikhi2017} might reduce computational effort while still
identifying high-quality configurations, particularly when extending the
approach to richer model classes or larger datasets. The benchmark
comparison includes selected NN and hybrid methods from the M4
Forecasting Competition, but it does not provide a systematic evaluation
of newer neural forecasting architectures. Finally, the empirical
analysis focused on monthly and quarterly frequencies; applying the
proposed ESN framework to other M4 frequencies (e.g., daily or hourly)
or to newer benchmark datasets would provide additional evidence on its
generality and limitations, especially in settings with complex multiple
seasonality or high-frequency noise.

In summary, the findings suggest that simple ESNs can be a useful
addition to the forecasting toolbox when many relatively short monthly
or quarterly series must be processed automatically. The ESN is
competitive with the strongest statistical benchmarks, particularly
under MASE, but ES-RNN achieves the best aggregate accuracy for both
frequencies. The evidence therefore supports empirical competitiveness
rather than uniform superiority. The conclusions are conditional on the
filtered M4 subsets, the first-order autoregressive ESN design, the fixed
preprocessing pipeline, and the selected hyperparameter grid.

\newpage

\section*{Appendix} \label{sec:appendix}

\subsection*{A.1 Data} \label{sec:appendix-data}

\begin{table}[hbt!]
\centering
\begin{tabular}{lrrrrrrrr}
\toprule
\multirow{3}{*}{\textbf{Application Field}}
& \multicolumn{4}{c}{\textbf{Parameter}} 
& \multicolumn{4}{c}{\textbf{Forecast}} \\
\cmidrule(l){2-5} \cmidrule(l){6-9}
& \multicolumn{2}{c}{\textbf{Monthly}} 
& \multicolumn{2}{c}{\textbf{Quarterly}} 
& \multicolumn{2}{c}{\textbf{Monthly}} 
& \multicolumn{2}{c}{\textbf{Quarterly}} \\
\cmidrule(l){2-3} \cmidrule(l){4-5}
\cmidrule(l){6-7} \cmidrule(l){8-9}
& $n$ & $\%$ & $n$ & $\%$ & $n$ & $\%$ & $n$ & $\%$ \\
\midrule
Demographic &   159 &   6.6 &    88 &   7.3 &   127 &   5.3 &    73 &   6.1 \\
Finance     &   693 &  28.9 &   210 &  17.5 &   696 &  29.0 &   214 &  17.8 \\
Industry    &   532 &  22.2 &   254 &  21.2 &   561 &  23.4 &   271 &  22.6 \\
Macro       &   413 &  17.2 &   251 &  20.9 &   396 &  16.5 &   223 &  18.6 \\
Micro       &   590 &  24.6 &   330 &  27.5 &   602 &  25.1 &   345 &  28.8 \\
Other       &    13 &   0.5 &    67 &   5.6 &    18 &   0.8 &    74 &   6.2 \\
\midrule
Total       & 2,400 & 100.0 & 1,200 & 100.0 & 2,400 & 100.0 & 1,200 & 100.0 \\
\bottomrule
\end{tabular}
\caption[Application-field frequencies by dataset]{Absolute ($n$) and relative ($\%$) frequencies by application field for the datasets \textit{Parameter} and \textit{Forecast} per frequency.}
\label{tab:dataset}
\end{table}

\noindent \textbf{\textit{Measuring strength of trend and strength of seasonality}}

To measure the strength of trend and strength of seasonality, the time
series is decomposed via STL (Seasonal and Trend decomposition using
LOESS) into the additive model \(y_{t} = T_{t} + S_{t} + R_{t}\), where
\(T_{t}\) is the trend-cycle component, \(S_{t}\) is the seasonal
component and \(R_{t}\) is the remainder component
\citep{Cleveland1990}. For time series with a strong trend, the
seasonally adjusted series should have much more variation than the
remainder component, and therefore
\(\operatorname{Var}(R_{t}) / \operatorname{Var}(T_{t} + R_{t})\) should
be relatively small. The two variances should be approximately the same
for time series with little or no trend. The strength of the trend can
be defined as

\begin{equation}
F_{T} = \max\left(0, 1 - \frac{\operatorname{Var}(R_{t})}{\operatorname{Var}(T_{t} + R_{t})}\right),
\label{eq:trend}
\end{equation}

\noindent which measures the trend's strength between 0 and 1. The
minimal possible value of \(F_{T}\) is set equal to zero as the variance
of the remainder can be larger than the variance of the seasonally
adjusted series.

The strength of seasonality can be defined as

\begin{equation}
F_{S} = \max\left(0, 1 - \frac{\operatorname{Var}(R_{t})}{\operatorname{Var}(S_{t} + R_{t})}\right),
\label{eq:season}
\end{equation}

\noindent similar to the trend strength, but for the detrended time
series instead of the seasonally adjusted data. A value for \(F_{S}\)
close to zero indicates almost no seasonality, while a value close to
one exhibits a strong seasonal pattern within the time series because
\(\operatorname{Var}(R_{t})\) will be much smaller than
\(\operatorname{Var}(S_{t} + R_{t})\). These measures are helpful when
analyzing a large number of time series to get an overview of the time
series characteristics \citep{Hyndman2021, feasts}.

\begin{table}[hbt!]
\centering
\begin{tabular}{llrrrrrrr}
\toprule
 \textbf{Metric} & \textbf{Dataset} & \textbf{Min.} & $\mathbf{Q_{1}}$ & \textbf{Mean} & \textbf{Median} & $\mathbf{Q_{3}}$ & \textbf{Max.} & \textbf{Std.} \\ 
\midrule
\multirow{3}{*}{Length} 
& Total & 42.000 & 82.000 & 216.300 & 202.000 & 306.000 & 2794.000 & 137.406 \\ 
& Parameter & 54.000 & 69.000 & 127.965 & 102.000 & 184.000 & 240.000 & 62.435 \\ 
& Forecast & 46.000 & 69.000 & 129.592 & 104.000 & 186.000 & 240.000 & 62.391 \\ 
 \midrule
\multirow{3}{*}{Season}
& Total & 0.000 & 0.192 & 0.415 & 0.326 & 0.634 & 0.999 & 0.277 \\ 
& Parameter & 0.000 & 0.174 & 0.339 & 0.268 & 0.469 & 0.994 & 0.234 \\ 
& Forecast & 0.000 & 0.170 & 0.341 & 0.263 & 0.478 & 0.994 & 0.242 \\ 
\midrule
\multirow{3}{*}{Trend}
& Total & 0.015 & 0.894 & 0.889 & 0.971 & 0.993 & 1.000 & 0.188 \\ 
& Parameter & 0.037 & 0.805 & 0.848 & 0.953 & 0.983 & 1.000 & 0.214 \\ 
& Forecast & 0.015 & 0.813 & 0.847 & 0.950 & 0.983 & 1.000 & 0.213 \\ 
\bottomrule
\end{tabular}
\caption[Monthly descriptive statistics]{Descriptive statistics of the \textbf{monthly dataset}, summarizing the number of observations, seasonal strength and trend strength per time series for the complete M4 dataset ($n = 48,000$) and the randomly sampled datasets \textit{Parameter} and \textit{Forecast} (each $n = 2,400$).}
\label{tab:descriptive-monthly}
\end{table}

\begin{table}[hbt!]
\centering
\begin{tabular}{llrrrrrrr}
\toprule
 \textbf{Metric} & \textbf{Dataset} & \textbf{Min.} & $\mathbf{Q_{1}}$ & \textbf{Mean} & \textbf{Median} & $\mathbf{Q_{3}}$ & \textbf{Max.} & \textbf{Std.} \\ 
\midrule
\multirow{3}{*}{Length}
& Total & 16.000 & 62.000 & 92.254 & 88.000 & 115.000 & 866.000 & 51.130 \\ 
& Parameter & 17.000 & 39.000 & 55.522 & 58.000 & 70.000 & 80.000 & 15.524 \\ 
& Forecast & 20.000 & 39.000 & 56.038 & 61.500 & 70.000 & 80.000 & 15.590 \\ 
 \midrule
\multirow{3}{*}{Season}
& Total & 0.000 & 0.148 & 0.335 & 0.248 & 0.454 & 0.999 & 0.259 \\ 
& Parameter & 0.000 & 0.138 & 0.312 & 0.222 & 0.394 & 0.995 & 0.250 \\ 
& Forecast & 0.000 & 0.131 & 0.304 & 0.221 & 0.391 & 0.995 & 0.245 \\ 
\midrule
\multirow{3}{*}{Trend}
& Total & 0.149 & 0.959 & 0.954 & 0.990 & 0.998 & 1.000 & 0.092 \\ 
& Parameter & 0.361 & 0.946 & 0.946 & 0.986 & 0.996 & 1.000 & 0.097 \\ 
& Forecast & 0.289 & 0.943 & 0.941 & 0.983 & 0.996 & 1.000 & 0.108 \\ 
\bottomrule
\end{tabular}
\caption[Quarterly descriptive statistics]{Descriptive statistics of the \textbf{quarterly dataset}, summarizing the number of observations, seasonal strength and trend strength per time series for the complete M4 dataset ($n = 24,000$) and the randomly sampled datasets \textit{Parameter} and \textit{Forecast} (each $n = 1,200$).}
\label{tab:descriptive-quarterly}
\end{table}

\newpage

\subsection*{A.2 Hyperparameter sweep} \label{sec:appendix-hyperparameter}

\noindent \textbf{\textit{Calculation of information criteria}}

For each candidate ridge penalty \(\lambda\), the fitted readout produces
the residual sum of squares

\begin{equation}
RSS_{\lambda}
=
\left\|
\mathbf{y}
-
\mathbf{X}
\widehat{\mathbf{W}}_{\lambda}^{\mathrm{out}}
\right\|_{2}^{2}.
\label{eq:rss-lambda}
\end{equation}

Let \(n\) denote the number of observations used to estimate the
readout after differencing, lag construction, and removal of the washout
period. The effective degrees-of-freedom \(df_{\lambda}\) are calculated
from the ridge hat matrix as defined in Equation \ref{eq:dof}. The
intercept is not penalized.

Assuming Gaussian readout residuals, the maximized log-likelihood used
in the implementation is

\begin{equation}
\ell_{\lambda}
=
-\frac{n}{2}
\left[
\log(2\pi)
+
1
+
\log\left(
\frac{RSS_{\lambda}}{n}
\right)
\right].
\label{eq:loglik-lambda}
\end{equation}

Based on this log-likelihood and the effective degrees-of-freedom, the
information criteria are calculated as

\begin{equation}
AIC_{\lambda}
=
-2\ell_{\lambda}
+
2df_{\lambda}.
\label{eq:aic-lambda}
\end{equation}

\begin{equation}
AICc_{\lambda}
=
AIC_{\lambda}
+
\frac{
2df_{\lambda}(df_{\lambda}+1)
}{
n-df_{\lambda}-1
}.
\label{eq:aicc-lambda}
\end{equation}

\begin{equation}
BIC_{\lambda}
=
-2\ell_{\lambda}
+
df_{\lambda}\log(n).
\label{eq:bic-lambda}
\end{equation}

\begin{equation}
HQC_{\lambda}
=
-2\ell_{\lambda}
+
2df_{\lambda}\log\!\left(\log(n)\right).
\label{eq:hqc-lambda}
\end{equation}

For each fixed ESN configuration and reservoir realization, these
criteria are evaluated for all randomly drawn candidate values of
\(\lambda\). The candidate producing the lowest value of the selected
information criterion is retained for forecasting.

\begin{sidewaystable}
\centering
\begin{tabular}{llrrrrrrrrrrrrrr}
\toprule
\multirow{2}{*}{\textbf{Par.}} & \multirow{2}{*}{\textbf{Value}} & \multicolumn{7}{c}{\textbf{MASE}} & \multicolumn{7}{c}{\textbf{sMAPE [\%]}} \\
\cmidrule(l){3-9} \cmidrule(l){10-16}
 & & Min. & $Q_1$ & Mean & Median & $Q_3$ & Max. & Std. & Min. & $Q_1$ & Mean & Median & $Q_3$ & Max. & Std. \\
\midrule
\multirow{10}{*}{$\alpha$} 
& 0.1 & 0.010 & 0.494 & 1.047 & 0.813 & 1.317 & 129.508 & 0.984 & 0.074 & 4.880 & 20.326 & 14.090 & 28.692 & 197.694 & 21.576 \\ 
& 0.2 & 0.014 & 0.490 & 1.034 & 0.807 & 1.308 & 117.064 & 1.014 & 0.070 & 4.761 & 20.375 & 14.035 & 28.418 & 200.000 & 22.137 \\ 
& 0.3 & 0.015 & 0.482 & 1.001 & 0.789 & 1.281 & 10.102 & 0.804 & 0.069 & 4.648 & 19.864 & 13.628 & 27.725 & 199.616 & 21.513 \\ 
& 0.4 & 0.012 & 0.471 & 0.973 & 0.771 & 1.251 & 14.790 & 0.777 & 0.071 & 4.553 & 19.040 & 13.323 & 26.820 & 199.206 & 20.057 \\ 
& 0.5 & 0.012 & 0.466 & 0.957 & 0.763 & 1.230 & 10.521 & 0.759 & 0.073 & 4.488 & 18.676 & 13.143 & 26.524 & 200.000 & 19.427 \\ 
& 0.6 & 0.007 & 0.462 & 0.944 & 0.756 & 1.211 & 10.011 & 0.746 & 0.073 & 4.434 & 18.452 & 13.053 & 26.274 & 198.338 & 19.109 \\ 
& 0.7 & 0.005 & 0.459 & 0.938 & 0.752 & 1.201 & 10.209 & 0.742 & 0.071 & 4.413 & 18.309 & 13.020 & 26.174 & 200.000 & 18.852 \\ 
& 0.8 & 0.008 & 0.457 & 0.935 & 0.748 & 1.196 & 17.255 & 0.745 & 0.070 & 4.398 & 18.240 & 12.918 & 26.048 & 200.000 & 18.782 \\ 
& 0.9 & 0.003 & 0.455 & 0.935 & 0.747 & 1.194 & 154.659 & 0.796 & 0.074 & 4.392 & 18.242 & 12.862 & 26.066 & 200.000 & 18.886 \\ 
& 1.0 & 0.005 & 0.454 & 0.937 & 0.748 & 1.200 & 36.473 & 0.757 & 0.073 & 4.384 & 18.312 & 12.847 & 26.057 & 200.000 & 19.020 \\ 
\midrule
\multirow{11}{*}{$\rho$}
& 0.2 & 0.005 & 0.460 & 0.969 & 0.759 & 1.239 & 10.586 & 0.808 & 0.104 & 4.468 & 18.334 & 13.065 & 26.200 & 189.054 & 18.632 \\ 
& 0.3 & 0.007 & 0.459 & 0.961 & 0.756 & 1.233 & 10.586 & 0.795 & 0.071 & 4.444 & 18.250 & 13.050 & 26.081 & 189.552 & 18.548 \\ 
& 0.4 & 0.007 & 0.457 & 0.954 & 0.753 & 1.225 & 10.621 & 0.782 & 0.069 & 4.428 & 18.198 & 13.032 & 26.028 & 189.901 & 18.514 \\ 
& 0.5 & 0.014 & 0.456 & 0.947 & 0.750 & 1.217 & 10.802 & 0.767 & 0.071 & 4.424 & 18.154 & 13.025 & 25.956 & 190.010 & 18.480 \\ 
& 0.6 & 0.003 & 0.455 & 0.940 & 0.748 & 1.206 & 11.095 & 0.754 & 0.070 & 4.417 & 18.117 & 13.007 & 25.986 & 189.932 & 18.470 \\ 
& 0.7 & 0.005 & 0.455 & 0.934 & 0.747 & 1.198 & 11.215 & 0.743 & 0.070 & 4.407 & 18.123 & 13.007 & 26.024 & 191.672 & 18.520 \\ 
& 0.8 & 0.015 & 0.456 & 0.932 & 0.748 & 1.194 & 11.157 & 0.736 & 0.071 & 4.422 & 18.247 & 13.056 & 26.167 & 193.542 & 18.730 \\ 
& 0.9 & 0.010 & 0.464 & 0.939 & 0.757 & 1.201 & 10.906 & 0.735 & 0.071 & 4.449 & 18.618 & 13.126 & 26.552 & 194.810 & 19.375 \\ 
& 1.0 & 0.010 & 0.479 & 0.967 & 0.778 & 1.232 & 117.064 & 0.867 & 0.072 & 4.562 & 19.545 & 13.387 & 27.456 & 198.901 & 21.019 \\ 
& 1.1 & 0.012 & 0.503 & 1.028 & 0.816 & 1.308 & 85.601 & 0.869 & 0.072 & 4.839 & 21.002 & 14.004 & 29.056 & 200.000 & 23.185 \\ 
& 1.2 & 0.011 & 0.525 & 1.099 & 0.856 & 1.386 & 154.659 & 1.078 & 0.070 & 5.036 & 22.231 & 14.644 & 30.612 & 200.000 & 24.736 \\ 
\midrule
\multirow{3}{*}{$\tau$}
& 0.2 & 0.007 & 0.469 & 0.974 & 0.773 & 1.251 & 10.588 & 0.784 & 0.070 & 4.513 & 19.145 & 13.237 & 27.125 & 200.000 & 20.201 \\ 
& 0.4 & 0.005 & 0.468 & 0.966 & 0.766 & 1.233 & 129.508 & 0.873 & 0.069 & 4.497 & 18.890 & 13.220 & 26.709 & 200.000 & 19.861 \\ 
& 0.6 & 0.003 & 0.468 & 0.970 & 0.768 & 1.238 & 154.659 & 0.797 & 0.070 & 4.582 & 18.917 & 13.319 & 26.663 & 200.000 & 19.911 \\ 
\midrule
\multirow{4}{*}{IC}
& AIC  & 0.005 & 0.471 & 0.977 & 0.773 & 1.247 & 154.659 & 0.840 & 0.069 & 4.574 & 19.116 & 13.328 & 26.940 & 200.000 & 20.212 \\ 
& AICc & 0.005 & 0.470 & 0.972 & 0.770 & 1.243 & 129.508 & 0.816 & 0.070 & 4.548 & 19.030 & 13.287 & 26.867 & 200.000 & 20.067 \\ 
& BIC  & 0.003 & 0.466 & 0.964 & 0.766 & 1.234 & 129.508 & 0.808 & 0.071 & 4.484 & 18.840 & 13.191 & 26.702 & 200.000 & 19.743 \\ 
& HQC  & 0.003 & 0.467 & 0.968 & 0.768 & 1.239 & 129.508 & 0.812 & 0.071 & 4.514 & 18.948 & 13.239 & 26.788 & 200.000 & 19.941 \\
\bottomrule
\end{tabular}
\caption[Monthly hyperparameter accuracy distributions]{Descriptive statistics, summarizing the distribution of out-of-sample forecast accuracy per hyperparameter for $n = 2,400$ \textbf{monthly time series}. The table lists the minimum, first quartile ($Q_1$), mean, median, third quartile ($Q_3$), maximum, and standard deviation for MASE and sMAPE (\%). The rows are grouped by hyperparameter and ordered by value.}
\label{tab:pars-monthly-dist}
\end{sidewaystable}

\begin{sidewaystable}
\centering
\begin{tabular}{llrrrrrrrrrrrrrr}
\toprule
\multirow{2}{*}{\textbf{Par.}} & \multirow{2}{*}{\textbf{Value}} & \multicolumn{7}{c}{\textbf{MASE}} & \multicolumn{7}{c}{\textbf{sMAPE [\%]}} \\
\cmidrule(l){3-9} \cmidrule(l){10-16}
 & & Min. & $Q_1$ & Mean & Median & $Q_3$ & Max. & Std. & Min. & $Q_1$ & Mean & Median & $Q_3$ & Max. & Std. \\
\midrule
\multirow{10}{*}{$\alpha$}
& 0.1 & 0.028 & 0.534 & 1.185 & 0.946 & 1.556 & 7.153 & 0.914 & 0.079 & 2.664 & 11.006 & 5.854 & 12.916 & 188.934 & 14.906 \\ 
& 0.2 & 0.037 & 0.535 & 1.187 & 0.954 & 1.578 & 7.309 & 0.908 & 0.113 & 2.690 & 11.179 & 5.902 & 13.035 & 200.000 & 15.663 \\ 
& 0.3 & 0.025 & 0.534 & 1.186 & 0.954 & 1.574 & 7.392 & 0.903 & 0.074 & 2.702 & 11.142 & 5.913 & 13.051 & 198.300 & 15.609 \\ 
& 0.4 & 0.026 & 0.529 & 1.177 & 0.948 & 1.560 & 7.876 & 0.894 & 0.073 & 2.683 & 11.070 & 5.897 & 12.847 & 200.000 & 15.591 \\ 
& 0.5 & 0.024 & 0.526 & 1.165 & 0.936 & 1.545 & 9.053 & 0.887 & 0.067 & 2.651 & 10.953 & 5.855 & 12.670 & 195.642 & 15.432 \\ 
& 0.6 & 0.024 & 0.520 & 1.152 & 0.923 & 1.525 & 7.390 & 0.877 & 0.067 & 2.606 & 10.834 & 5.779 & 12.517 & 195.763 & 15.293 \\ 
& 0.7 & 0.024 & 0.515 & 1.140 & 0.909 & 1.507 & 8.060 & 0.868 & 0.067 & 2.549 & 10.729 & 5.740 & 12.322 & 191.572 & 15.166 \\ 
& 0.8 & 0.024 & 0.509 & 1.127 & 0.900 & 1.484 & 8.334 & 0.859 & 0.067 & 2.519 & 10.627 & 5.696 & 12.177 & 200.000 & 15.083 \\ 
& 0.9 & 0.015 & 0.504 & 1.118 & 0.894 & 1.466 & 10.447 & 0.854 & 0.058 & 2.480 & 10.579 & 5.654 & 12.116 & 190.702 & 15.109 \\ 
& 1.0 & 0.012 & 0.502 & 1.112 & 0.890 & 1.457 & 9.013 & 0.852 & 0.046 & 2.450 & 10.525 & 5.613 & 12.085 & 198.089 & 15.027 \\ 
\midrule
\multirow{11}{*}{$\rho$}
& 0.2 & 0.026 & 0.499 & 1.124 & 0.896 & 1.485 & 7.423 & 0.865 & 0.074 & 2.480 & 10.492 & 5.678 & 12.081 & 187.019 & 14.648 \\ 
& 0.3 & 0.028 & 0.502 & 1.125 & 0.895 & 1.484 & 7.416 & 0.863 & 0.077 & 2.495 & 10.497 & 5.679 & 12.089 & 187.877 & 14.608 \\ 
& 0.4 & 0.025 & 0.504 & 1.127 & 0.898 & 1.488 & 7.405 & 0.863 & 0.073 & 2.510 & 10.517 & 5.686 & 12.145 & 189.066 & 14.607 \\ 
& 0.5 & 0.024 & 0.507 & 1.130 & 0.903 & 1.491 & 7.390 & 0.865 & 0.067 & 2.520 & 10.555 & 5.697 & 12.220 & 188.848 & 14.644 \\ 
& 0.6 & 0.024 & 0.510 & 1.135 & 0.909 & 1.495 & 7.378 & 0.868 & 0.067 & 2.530 & 10.611 & 5.723 & 12.271 & 188.074 & 14.720 \\ 
& 0.7 & 0.025 & 0.514 & 1.141 & 0.915 & 1.506 & 7.367 & 0.871 & 0.077 & 2.560 & 10.686 & 5.745 & 12.382 & 187.595 & 14.841 \\ 
& 0.8 & 0.014 & 0.520 & 1.149 & 0.923 & 1.518 & 7.425 & 0.874 & 0.054 & 2.609 & 10.789 & 5.757 & 12.498 & 187.124 & 15.032 \\ 
& 0.9 & 0.012 & 0.529 & 1.161 & 0.934 & 1.533 & 9.017 & 0.880 & 0.046 & 2.639 & 10.933 & 5.823 & 12.658 & 192.977 & 15.331 \\ 
& 1.0 & 0.016 & 0.537 & 1.177 & 0.950 & 1.554 & 9.013 & 0.888 & 0.064 & 2.680 & 11.146 & 5.909 & 12.866 & 198.870 & 15.747 \\ 
& 1.1 & 0.026 & 0.550 & 1.204 & 0.966 & 1.589 & 10.447 & 0.914 & 0.073 & 2.755 & 11.495 & 6.022 & 13.278 & 200.000 & 16.502 \\ 
& 1.2 & 0.026 & 0.557 & 1.228 & 0.978 & 1.620 & 9.771 & 0.943 & 0.074 & 2.794 & 11.788 & 6.100 & 13.509 & 198.803 & 17.212 \\ 
\midrule
\multirow{3}{*}{$\tau$}
& 0.2 & 0.024 & 0.520 & 1.153 & 0.924 & 1.518 & 9.017 & 0.884 & 0.067 & 2.599 & 10.768 & 5.825 & 12.474 & 188.014 & 14.957 \\ 
& 0.4 & 0.012 & 0.519 & 1.150 & 0.923 & 1.519 & 7.890 & 0.876 & 0.046 & 2.580 & 10.837 & 5.749 & 12.567 & 200.000 & 15.221 \\ 
& 0.6 & 0.016 & 0.524 & 1.161 & 0.930 & 1.537 & 10.447 & 0.887 & 0.064 & 2.600 & 10.988 & 5.801 & 12.641 & 200.000 & 15.687 \\ 
\midrule
\multirow{4}{*}{IC}
& AIC & 0.012 & 0.524 & 1.163 & 0.931 & 1.537 & 10.447 & 0.890 & 0.046 & 2.617 & 10.956 & 5.823 & 12.673 & 200.000 & 15.475 \\ 
& AICc & 0.012 & 0.522 & 1.156 & 0.926 & 1.529 & 9.053 & 0.883 & 0.046 & 2.599 & 10.882 & 5.797 & 12.582 & 200.000 & 15.331 \\ 
& BIC & 0.012 & 0.518 & 1.146 & 0.920 & 1.511 & 8.286 & 0.874 & 0.046 & 2.562 & 10.766 & 5.756 & 12.457 & 200.000 & 15.076 \\ 
& HQC & 0.012 & 0.521 & 1.154 & 0.925 & 1.524 & 9.053 & 0.882 & 0.046 & 2.592 & 10.853 & 5.781 & 12.548 & 200.000 & 15.281 \\ 

\bottomrule
\end{tabular}
\caption[Quarterly hyperparameter accuracy distributions]{Descriptive statistics, summarizing the distribution of out-of-sample forecast accuracy per hyperparameter for $n = 1,200$ \textbf{quarterly time series}. The table lists the minimum, first quartile ($Q_1$), mean, median, third quartile ($Q_3$), maximum, and standard deviation for MASE and sMAPE (\%). The rows are grouped by hyperparameter and ordered by value.}
\label{tab:pars-quarterly-dist}
\end{sidewaystable}

\subsection*{A.3 Random initialization} \label{sec:appendix-random}

\begin{sidewaystable}
\centering
\begin{tabular}{rrrrrrrrrrrrrrr}
\toprule
\multirow{2}{*}{\textbf{Seed}} &
\multicolumn{7}{c}{\textbf{MASE}} &
\multicolumn{7}{c}{\textbf{sMAPE [\%]}} \\
\cmidrule(l){2-8} \cmidrule(l){9-15}
& Min. & $Q_1$ & Mean & Median & $Q_3$ & Max. & Std.
& Min. & $Q_1$ & Mean & Median & $Q_3$ & Max. & Std. \\
\midrule
24 & 0.011 & 0.440 & 0.893 & 0.721 & 1.145 & 7.138 & 0.697 & 0.154 & 4.143 & 17.859 & 12.560 & 25.855 & 187.578 & 18.605 \\ 
42 & 0.023 & 0.425 & 0.878 & 0.712 & 1.137 & 7.167 & 0.680 & 0.078 & 4.097 & 17.592 & 12.406 & 25.324 & 187.273 & 18.287 \\ 
50 & 0.013 & 0.432 & 0.886 & 0.725 & 1.141 & 7.166 & 0.684 & 0.119 & 4.223 & 17.751 & 12.681 & 25.428 & 186.516 & 18.503 \\ 
75 & 0.022 & 0.439 & 0.885 & 0.710 & 1.148 & 7.192 & 0.673 & 0.095 & 4.216 & 17.819 & 12.735 & 25.621 & 187.880 & 18.569 \\ 
129 & 0.022 & 0.436 & 0.887 & 0.716 & 1.138 & 7.764 & 0.690 & 0.082 & 4.231 & 17.741 & 12.542 & 26.047 & 187.984 & 18.472 \\ 
147 & 0.016 & 0.435 & 0.886 & 0.716 & 1.133 & 7.110 & 0.688 & 0.119 & 4.234 & 17.687 & 12.599 & 25.795 & 187.864 & 18.232 \\ 
154 & 0.030 & 0.440 & 0.888 & 0.723 & 1.141 & 7.176 & 0.677 & 0.083 & 4.232 & 17.863 & 12.727 & 25.820 & 188.170 & 18.489 \\ 
166 & 0.016 & 0.435 & 0.894 & 0.722 & 1.146 & 7.119 & 0.692 & 0.091 & 4.179 & 17.908 & 12.625 & 25.950 & 187.371 & 18.809 \\ 
213 & 0.034 & 0.436 & 0.886 & 0.726 & 1.133 & 7.192 & 0.680 & 0.115 & 4.173 & 17.779 & 12.854 & 25.919 & 180.621 & 18.261 \\ 
229 & 0.025 & 0.433 & 0.885 & 0.707 & 1.131 & 7.149 & 0.693 & 0.080 & 4.195 & 17.750 & 12.498 & 25.503 & 182.495 & 18.462 \\ 
284 & 0.014 & 0.443 & 0.886 & 0.725 & 1.145 & 7.163 & 0.679 & 0.123 & 4.120 & 17.888 & 12.848 & 25.852 & 189.189 & 18.537 \\ 
298 & 0.017 & 0.442 & 0.889 & 0.723 & 1.138 & 7.142 & 0.676 & 0.088 & 4.135 & 17.868 & 12.708 & 26.047 & 191.852 & 18.392 \\ 
304 & 0.015 & 0.441 & 0.891 & 0.714 & 1.137 & 7.179 & 0.686 & 0.111 & 4.265 & 17.814 & 12.616 & 25.904 & 189.496 & 18.536 \\ 
322 & 0.023 & 0.441 & 0.891 & 0.721 & 1.130 & 7.172 & 0.689 & 0.086 & 4.145 & 17.850 & 12.932 & 25.592 & 187.978 & 18.521 \\ 
357 & 0.023 & 0.438 & 0.894 & 0.714 & 1.153 & 7.572 & 0.698 & 0.079 & 4.184 & 17.803 & 12.636 & 25.930 & 188.524 & 18.465 \\ 
411 & 0.022 & 0.437 & 0.890 & 0.727 & 1.141 & 8.483 & 0.694 & 0.083 & 4.178 & 17.865 & 12.698 & 25.897 & 187.261 & 18.758 \\ 
518 & 0.018 & 0.434 & 0.886 & 0.716 & 1.132 & 7.171 & 0.684 & 0.099 & 4.219 & 17.728 & 12.508 & 25.713 & 188.676 & 18.477 \\ 
533 & 0.022 & 0.432 & 0.885 & 0.720 & 1.135 & 7.148 & 0.684 & 0.170 & 4.103 & 17.693 & 12.878 & 25.785 & 189.458 & 18.351 \\ 
562 & 0.031 & 0.433 & 0.885 & 0.720 & 1.144 & 7.170 & 0.680 & 0.085 & 4.223 & 17.692 & 12.680 & 25.619 & 187.370 & 18.412 \\ 
602 & 0.026 & 0.439 & 0.891 & 0.725 & 1.134 & 7.181 & 0.686 & 0.097 & 4.170 & 17.835 & 12.726 & 25.907 & 185.723 & 18.566 \\ 
622 & 0.014 & 0.443 & 0.893 & 0.717 & 1.145 & 7.125 & 0.694 & 0.160 & 4.137 & 17.819 & 12.495 & 25.874 & 187.931 & 18.554 \\ 
623 & 0.021 & 0.439 & 0.894 & 0.725 & 1.145 & 7.207 & 0.690 & 0.132 & 4.173 & 17.863 & 12.701 & 26.142 & 187.672 & 18.430 \\ 
635 & 0.021 & 0.440 & 0.894 & 0.729 & 1.153 & 7.908 & 0.694 & 0.154 & 4.183 & 17.877 & 12.488 & 26.006 & 187.202 & 18.591 \\ 
840 & 0.021 & 0.442 & 0.907 & 0.717 & 1.133 & 60.403 & 1.387 & 0.132 & 4.211 & 17.641 & 12.627 & 25.551 & 187.959 & 18.341 \\ 
880 & 0.027 & 0.436 & 0.889 & 0.719 & 1.134 & 8.099 & 0.694 & 0.142 & 4.231 & 17.767 & 12.886 & 25.825 & 188.482 & 18.439 \\ 
883 & 0.011 & 0.439 & 0.892 & 0.730 & 1.147 & 7.187 & 0.679 & 0.100 & 4.142 & 17.898 & 13.074 & 25.815 & 182.699 & 18.445 \\ 
900 & 0.035 & 0.435 & 0.880 & 0.713 & 1.129 & 7.193 & 0.679 & 0.097 & 4.142 & 17.699 & 12.450 & 25.387 & 183.248 & 18.482 \\ 
933 & 0.022 & 0.439 & 0.889 & 0.726 & 1.144 & 7.156 & 0.680 & 0.127 & 4.127 & 17.815 & 12.743 & 25.755 & 188.500 & 18.539 \\ 
985 & 0.016 & 0.438 & 0.886 & 0.723 & 1.131 & 7.195 & 0.678 & 0.189 & 4.181 & 17.673 & 12.664 & 25.555 & 188.326 & 18.257 \\ 
998 & 0.025 & 0.436 & 0.889 & 0.717 & 1.151 & 7.175 & 0.693 & 0.084 & 4.232 & 17.754 & 12.669 & 25.738 & 187.699 & 18.416 \\ 
\bottomrule
\end{tabular}
\caption[Monthly accuracy distributions across random initializations]{Descriptive statistics summarizing the distribution of out-of-sample forecast accuracy across $n = 2,400$ \textbf{monthly time series} for each random initialization of the ESN. The table reports the minimum, first quartile ($Q_1$), mean, median, third quartile ($Q_3$), maximum, and standard deviation of MASE and sMAPE (\%). The rows are ordered by the random seed.}
\label{tab:rand-monthly-dist}
\end{sidewaystable}

\begin{sidewaystable}
\centering
\begin{tabular}{rrrrrrrrrrrrrrr}
\toprule
\multirow{2}{*}{\textbf{Seed}} &
\multicolumn{7}{c}{\textbf{MASE}} &
\multicolumn{7}{c}{\textbf{sMAPE [\%]}} \\
\cmidrule(l){2-8} \cmidrule(l){9-15}
& Min. & $Q_1$ & Mean & Median & $Q_3$ & Max. & Std.
& Min. & $Q_1$ & Mean & Median & $Q_3$ & Max. & Std. \\
\midrule
24 & 0.029 & 0.484 & 1.084 & 0.868 & 1.423 & 7.261 & 0.834 & 0.116 & 2.376 & 10.261 & 5.517 & 11.924 & 186.042 & 14.674 \\ 
42 & 0.030 & 0.488 & 1.078 & 0.875 & 1.417 & 7.285 & 0.828 & 0.155 & 2.357 & 10.243 & 5.491 & 11.796 & 185.617 & 14.688 \\ 
50 & 0.032 & 0.495 & 1.085 & 0.866 & 1.417 & 7.249 & 0.833 & 0.157 & 2.367 & 10.311 & 5.508 & 11.830 & 186.203 & 14.739 \\ 
75 & 0.032 & 0.493 & 1.086 & 0.866 & 1.427 & 7.301 & 0.830 & 0.092 & 2.375 & 10.294 & 5.483 & 11.843 & 185.597 & 14.687 \\ 
129 & 0.030 & 0.492 & 1.081 & 0.864 & 1.423 & 7.260 & 0.837 & 0.145 & 2.360 & 10.254 & 5.476 & 11.793 & 186.114 & 14.736 \\ 
147 & 0.031 & 0.488 & 1.083 & 0.875 & 1.427 & 7.274 & 0.824 & 0.143 & 2.390 & 10.277 & 5.493 & 11.751 & 185.585 & 14.684 \\ 
154 & 0.031 & 0.486 & 1.082 & 0.866 & 1.425 & 7.271 & 0.838 & 0.162 & 2.342 & 10.252 & 5.578 & 11.969 & 185.948 & 14.669 \\ 
166 & 0.032 & 0.494 & 1.085 & 0.866 & 1.428 & 7.288 & 0.831 & 0.160 & 2.372 & 10.299 & 5.510 & 11.853 & 184.373 & 14.698 \\ 
213 & 0.032 & 0.495 & 1.086 & 0.868 & 1.422 & 7.299 & 0.828 & 0.124 & 2.426 & 10.282 & 5.489 & 11.798 & 185.334 & 14.649 \\ 
229 & 0.031 & 0.488 & 1.085 & 0.872 & 1.424 & 7.263 & 0.831 & 0.115 & 2.404 & 10.263 & 5.530 & 11.958 & 185.324 & 14.641 \\ 
284 & 0.031 & 0.492 & 1.088 & 0.884 & 1.430 & 7.244 & 0.833 & 0.151 & 2.421 & 10.311 & 5.523 & 11.890 & 185.924 & 14.686 \\ 
298 & 0.031 & 0.488 & 1.086 & 0.871 & 1.429 & 7.276 & 0.828 & 0.157 & 2.381 & 10.324 & 5.556 & 11.962 & 185.392 & 14.707 \\ 
304 & 0.032 & 0.495 & 1.084 & 0.869 & 1.424 & 7.267 & 0.836 & 0.140 & 2.366 & 10.262 & 5.478 & 11.872 & 185.856 & 14.667 \\ 
322 & 0.028 & 0.488 & 1.079 & 0.866 & 1.431 & 7.291 & 0.828 & 0.078 & 2.380 & 10.253 & 5.480 & 11.742 & 185.909 & 14.691 \\ 
357 & 0.031 & 0.490 & 1.084 & 0.867 & 1.421 & 7.263 & 0.835 & 0.125 & 2.365 & 10.264 & 5.515 & 11.720 & 185.168 & 14.705 \\ 
411 & 0.031 & 0.484 & 1.083 & 0.877 & 1.425 & 7.260 & 0.829 & 0.155 & 2.361 & 10.274 & 5.505 & 11.786 & 186.554 & 14.703 \\ 
518 & 0.030 & 0.492 & 1.089 & 0.871 & 1.428 & 7.250 & 0.837 & 0.119 & 2.428 & 10.273 & 5.503 & 11.856 & 186.591 & 14.680 \\ 
533 & 0.031 & 0.486 & 1.079 & 0.863 & 1.416 & 7.280 & 0.831 & 0.118 & 2.364 & 10.252 & 5.505 & 11.774 & 185.583 & 14.673 \\ 
562 & 0.032 & 0.492 & 1.080 & 0.869 & 1.425 & 7.262 & 0.829 & 0.127 & 2.375 & 10.249 & 5.533 & 11.692 & 185.487 & 14.657 \\ 
602 & 0.032 & 0.490 & 1.082 & 0.860 & 1.437 & 7.286 & 0.829 & 0.118 & 2.366 & 10.281 & 5.463 & 11.842 & 186.318 & 14.731 \\ 
622 & 0.029 & 0.486 & 1.078 & 0.871 & 1.406 & 7.251 & 0.830 & 0.086 & 2.372 & 10.241 & 5.488 & 11.794 & 185.401 & 14.691 \\ 
623 & 0.032 & 0.490 & 1.082 & 0.864 & 1.422 & 7.234 & 0.831 & 0.165 & 2.384 & 10.227 & 5.462 & 11.676 & 186.011 & 14.652 \\ 
635 & 0.032 & 0.491 & 1.079 & 0.867 & 1.408 & 7.283 & 0.829 & 0.150 & 2.303 & 10.253 & 5.500 & 11.596 & 186.178 & 14.726 \\ 
840 & 0.031 & 0.480 & 1.083 & 0.874 & 1.419 & 7.288 & 0.833 & 0.133 & 2.355 & 10.264 & 5.560 & 12.004 & 185.581 & 14.673 \\ 
880 & 0.032 & 0.491 & 1.086 & 0.871 & 1.432 & 7.271 & 0.829 & 0.116 & 2.397 & 10.308 & 5.484 & 11.733 & 186.141 & 14.726 \\ 
883 & 0.031 & 0.488 & 1.086 & 0.864 & 1.424 & 7.291 & 0.834 & 0.139 & 2.453 & 10.287 & 5.478 & 11.878 & 185.533 & 14.684 \\ 
900 & 0.031 & 0.491 & 1.086 & 0.870 & 1.430 & 7.262 & 0.833 & 0.150 & 2.453 & 10.303 & 5.459 & 11.945 & 185.924 & 14.705 \\ 
933 & 0.032 & 0.486 & 1.080 & 0.867 & 1.420 & 7.268 & 0.825 & 0.157 & 2.340 & 10.249 & 5.479 & 11.715 & 186.079 & 14.676 \\ 
985 & 0.030 & 0.492 & 1.082 & 0.862 & 1.428 & 7.281 & 0.828 & 0.084 & 2.384 & 10.253 & 5.472 & 11.733 & 184.770 & 14.644 \\ 
998 & 0.031 & 0.491 & 1.088 & 0.871 & 1.428 & 7.254 & 0.833 & 0.157 & 2.369 & 10.284 & 5.555 & 11.858 & 185.502 & 14.666 \\ 
\bottomrule
\end{tabular}
\caption[Quarterly accuracy distributions across random initializations]{Descriptive statistics summarizing the distribution of out-of-sample forecast accuracy across $n = 1,200$ \textbf{quarterly time series} for each random initialization of the ESN. The table reports the minimum, first quartile ($Q_1$), mean, median, third quartile ($Q_3$), maximum, and standard deviation of MASE and sMAPE (\%). The rows are ordered by the random seed.}
\label{tab:rand-quarterly-dist}
\end{sidewaystable}

\newpage

\subsection*{A.4 Benchmark analysis} \label{sec:appendix-benchmark}

\begin{table}[hbt!]
\centering
\begin{tabular}{rlrrrrrrr}
\toprule
\textbf{Rank} & \textbf{Model} & \textbf{Min.} &
$\mathbf{Q_{1}}$ & \textbf{Mean} & \textbf{Median} &
$\mathbf{Q_{3}}$ & \textbf{Max.} & \textbf{Std.} \\
\midrule
1  & ES-RNN   & 0.041 & 0.427 & 0.870 & 0.679 & 1.040 & 18.354 & 0.932 \\
2  & ARIMA    & 0.014 & 0.451 & 0.897 & 0.733 & 1.093 & 14.453 & 0.794 \\
\rowcolor{light-gray}
3  & ESN-0650 & 0.037 & 0.438 & 0.898 & 0.723 & 1.117 & 13.837 & 0.794 \\
4  & TBATS    & 0.008 & 0.449 & 0.899 & 0.718 & 1.112 & 15.235 & 0.809 \\
5  & ETS      & 0.000 & 0.458 & 0.905 & 0.713 & 1.129 & 12.496 & 0.775 \\
6  & THETA    & 0.050 & 0.437 & 0.911 & 0.724 & 1.131 & 15.275 & 0.823 \\
7  & NAIVE2   & 0.024 & 0.500 & 0.999 & 0.803 & 1.254 & 15.548 & 0.839 \\
8  & DRIFT    & 0.036 & 0.470 & 1.101 & 0.835 & 1.396 & 17.578 & 1.061 \\
9  & NAIVE    & 0.000 & 0.527 & 1.110 & 0.859 & 1.384 & 16.099 & 1.029 \\
10 & SNAIVE   & 0.000 & 0.624 & 1.195 & 0.967 & 1.484 & 20.416 & 1.024 \\
11 & RNN      & 0.060 & 0.741 & 1.514 & 1.145 & 1.908 & 19.448 & 1.249 \\
12 & MLP      & 0.081 & 0.726 & 1.544 & 1.147 & 1.954 & 18.923 & 1.338 \\
13 & MEAN     & 0.123 & 1.115 & 2.893 & 1.992 & 3.565 & 24.049 & 2.656 \\
\bottomrule
\end{tabular}
\caption[Monthly MASE by forecast model]{\textbf{Monthly MASE:}
Descriptive statistics summarizing the distribution of MASE per
forecasting model for the 2,400 monthly time series in the dataset
\textit{Forecast}. The models are ranked according to mean MASE from low
to high, and the ESN is shaded in light gray.}
\label{tab:mase-monthly}
\end{table}

\begin{table}[hbt!]
\centering
\begin{tabular}{rlrrrrrrr}
\toprule
\textbf{Rank} & \textbf{Model} & \textbf{Min.} &
$\mathbf{Q_{1}}$ & \textbf{Mean} & \textbf{Median} &
$\mathbf{Q_{3}}$ & \textbf{Max.} & \textbf{Std.} \\
\midrule
1  & ES-RNN   & 0.159 & 4.000 & 15.746 & 11.579 & 22.668 & 121.376 & 15.336 \\
2  & THETA    & 0.179 & 4.294 & 16.638 & 11.666 & 24.675 & 170.829 & 16.136 \\
3  & TBATS    & 0.075 & 4.110 & 16.686 & 11.845 & 24.335 & 140.503 & 16.228 \\
4  & ETS      & 0.000 & 4.158 & 17.257 & 11.696 & 25.118 & 155.392 & 17.360 \\
5  & ARIMA    & 0.077 & 4.103 & 17.334 & 11.845 & 24.976 & 129.936 & 17.383 \\
\rowcolor{light-gray}
6  & ESN-0650 & 0.257 & 4.215 & 17.463 & 12.109 & 25.025 & 177.725 & 17.706 \\
7  & NAIVE2   & 0.116 & 4.596 & 18.635 & 13.097 & 27.012 & 138.387 & 18.152 \\
8  & NAIVE    & 0.000 & 4.908 & 19.386 & 14.387 & 28.547 & 138.387 & 18.117 \\
9  & DRIFT    & 0.292 & 4.981 & 19.716 & 13.263 & 27.999 & 194.421 & 20.647 \\
10 & SNAIVE   & 0.000 & 6.471 & 20.642 & 15.602 & 29.903 & 137.976 & 18.114 \\
11 & MLP      & 0.271 & 8.038 & 30.791 & 18.725 & 38.992 & 200.000 & 34.501 \\
12 & RNN      & 0.363 & 7.956 & 32.041 & 18.959 & 40.357 & 200.000 & 37.554 \\
13 & MEAN     & 0.531 & 16.727 & 37.797 & 33.115 & 55.014 & 174.643 & 26.168 \\
\bottomrule
\end{tabular}
\caption[Monthly sMAPE by forecast model]{\textbf{Monthly sMAPE:}
Descriptive statistics summarizing the distribution of sMAPE per
forecasting model for the 2,400 monthly time series in the dataset
\textit{Forecast}. The models are ranked according to mean sMAPE from
low to high, and the ESN is shaded in light gray.}
\label{tab:smape-monthly}
\end{table}

\begin{table}[hbt!]
\centering
\begin{tabular}{rlrrrrrrr}
\toprule
\textbf{Rank} & \textbf{Model} & \textbf{Min.} &
$\mathbf{Q_{1}}$ & \textbf{Mean} & \textbf{Median} &
$\mathbf{Q_{3}}$ & \textbf{Max.} & \textbf{Std.} \\
\midrule
1  & ES-RNN   & 0.037 & 0.493 & 1.097 & 0.813 & 1.426 & 9.634 & 0.962 \\
\rowcolor{light-gray}
2  & ESN-0306 & 0.048 & 0.504 & 1.111 & 0.875 & 1.428 & 9.049 & 0.930 \\
3  & ETS      & 0.069 & 0.495 & 1.121 & 0.868 & 1.446 & 9.439 & 0.944 \\
4  & ARIMA    & 0.055 & 0.508 & 1.139 & 0.872 & 1.436 & 8.946 & 0.969 \\
5  & TBATS    & 0.055 & 0.524 & 1.160 & 0.879 & 1.449 & 9.522 & 1.014 \\
6  & THETA    & 0.043 & 0.550 & 1.171 & 0.922 & 1.496 & 9.048 & 0.948 \\
7  & DRIFT    & 0.058 & 0.529 & 1.192 & 0.908 & 1.523 & 9.885 & 1.041 \\
8  & NAIVE2   & 0.074 & 0.614 & 1.286 & 1.027 & 1.655 & 9.134 & 0.989 \\
9  & NAIVE    & 0.070 & 0.609 & 1.347 & 1.082 & 1.750 & 9.573 & 1.079 \\
10 & SNAIVE   & 0.097 & 0.746 & 1.522 & 1.281 & 1.977 & 9.850 & 1.108 \\
11 & MLP      & 0.059 & 0.825 & 1.935 & 1.454 & 2.593 & 42.974 & 1.917 \\
12 & RNN      & 0.073 & 0.823 & 1.995 & 1.485 & 2.584 & 99.922 & 3.224 \\
13 & MEAN     & 0.053 & 1.436 & 4.240 & 3.521 & 6.267 & 18.407 & 3.195 \\
\bottomrule
\end{tabular}
\caption[Quarterly MASE by forecast model]{\textbf{Quarterly MASE:}
Descriptive statistics summarizing the distribution of MASE per
forecasting model for the 1,200 quarterly time series in the dataset
\textit{Forecast}. The models are ranked according to mean MASE from low
to high, and the ESN is shaded in light gray.}
\label{tab:mase-quarterly}
\end{table}

\begin{table}[hbt!]
\centering
\begin{tabular}{rlrrrrrrr}
\toprule
\textbf{Rank} & \textbf{Model} & \textbf{Min.} &
$\mathbf{Q_{1}}$ & \textbf{Mean} & \textbf{Median} &
$\mathbf{Q_{3}}$ & \textbf{Max.} & \textbf{Std.} \\
\midrule
1  & ES-RNN   & 0.098 & 2.670 & 10.502 & 5.344 & 11.535 & 154.973 & 14.438 \\
2  & ETS      & 0.156 & 2.685 & 10.689 & 5.805 & 12.186 & 166.740 & 14.342 \\
3  & TBATS    & 0.128 & 2.743 & 10.832 & 6.028 & 12.629 & 167.089 & 14.175 \\
4  & ARIMA    & 0.147 & 2.758 & 10.917 & 5.875 & 12.839 & 129.042 & 13.953 \\
5  & THETA    & 0.117 & 3.076 & 10.932 & 6.186 & 12.434 & 164.638 & 14.528 \\
\rowcolor{light-gray}
6  & ESN-0306 & 0.177 & 2.590 & 11.023 & 5.701 & 12.272 & 168.859 & 15.533 \\
7  & NAIVE2   & 0.332 & 3.641 & 11.519 & 6.697 & 12.771 & 167.510 & 14.879 \\
8  & DRIFT    & 0.213 & 2.645 & 11.706 & 6.005 & 13.125 & 167.840 & 16.796 \\
9  & NAIVE    & 0.332 & 3.652 & 11.888 & 7.145 & 13.402 & 167.842 & 15.071 \\
10 & SNAIVE   & 0.359 & 4.599 & 13.185 & 8.194 & 15.556 & 153.666 & 14.840 \\
11 & MLP      & 0.133 & 4.499 & 17.505 & 9.866 & 21.747 & 200.000 & 21.734 \\
12 & RNN      & 0.182 & 4.559 & 17.584 & 10.048 & 21.649 & 189.089 & 21.700 \\
13 & MEAN     & 0.654 & 14.176 & 30.522 & 25.888 & 41.340 & 140.623 & 22.379 \\
\bottomrule
\end{tabular}
\caption[Quarterly sMAPE by forecast model]{\textbf{Quarterly sMAPE:}
Descriptive statistics summarizing the distribution of sMAPE per
forecasting model for the 1,200 quarterly time series in the dataset
\textit{Forecast}. The models are ranked according to mean sMAPE from
low to high, and the ESN is shaded in light gray.}
\label{tab:smape-quarterly}
\end{table}

\begin{table}[hbt!]
\centering
\begin{tabular}{lrrl}
\toprule
\textbf{Model} &
\textbf{Median difference} &
\textbf{Adjusted \boldmath{$p$}-value} &
\textbf{Decision} \\
\midrule
NAIVE  & -0.106 & $<0.001$ & ESN better \\
NAIVE2 & -0.059 & $<0.001$ & ESN better \\
SNAIVE & -0.175 & $<0.001$ & ESN better \\
MEAN   & -1.223 & $<0.001$ & ESN better \\
DRIFT  & -0.043 & $<0.001$ & ESN better \\
ARIMA  &  0.006 & 1.000    & Not significant \\
ETS    &  0.006 & 1.000    & Not significant \\
TBATS  &  0.008 & 1.000    & Not significant \\
THETA  &  0.012 & 1.000    & Not significant \\
RNN    & -0.361 & $<0.001$ & ESN better \\
MLP    & -0.338 & $<0.001$ & ESN better \\
ES-RNN &  0.032 & $<0.001$ & Benchmark better \\
\bottomrule
\end{tabular}
\caption[Pairwise MASE comparisons with ESN for monthly series]{Pairwise comparisons of MASE between the ESN and the benchmark models for the monthly series. Median differences are defined as $\mathrm{MASE}_{\mathrm{ESN}}-\mathrm{MASE}_{\mathrm{Benchmark}}$, such that negative values favor the ESN. Adjusted $p$-values are obtained from paired Wilcoxon signed-rank tests with Holm correction across all model comparisons. Results classified as ``ESN better'' or ``Benchmark better'' indicate significantly lower MASE for the corresponding model at the 5\% level.}
\label{tab:wilcoxon-monthly}
\end{table}

\begin{table}[hbt!]
\centering
\begin{tabular}{lrrl}
\toprule
\textbf{Model} &
\textbf{Median difference} &
\textbf{Adjusted \boldmath{$p$}-value} &
\textbf{Decision} \\
\midrule
NAIVE  & -0.145 & $<0.001$ & ESN better \\
NAIVE2 & -0.113 & $<0.001$ & ESN better \\
SNAIVE & -0.281 & $<0.001$ & ESN better \\
MEAN   & -2.506 & $<0.001$ & ESN better \\
DRIFT  & -0.011 & $<0.001$ & ESN better \\
ARIMA  &  0.005 & 1.000    & Not significant \\
ETS    & -0.006 & 1.000    & Not significant \\
TBATS  & -0.005 & 0.628    & Not significant \\
THETA  & -0.033 & $<0.001$ & ESN better \\
RNN    & -0.444 & $<0.001$ & ESN better \\
MLP    & -0.419 & $<0.001$ & ESN better \\
ES-RNN &  0.004 & 1.000    & Not significant \\
\bottomrule
\end{tabular}

\caption[Pairwise MASE comparisons with ESN for quarterly series]{Pairwise comparisons of MASE between the ESN and the benchmark models for the quarterly series. Median differences are defined as $\mathrm{MASE}_{\mathrm{ESN}}-\mathrm{MASE}_{\mathrm{Benchmark}}$, such that negative values favor the ESN. Adjusted $p$-values are obtained from paired Wilcoxon signed-rank tests with Holm correction across all model comparisons. Results classified as ``ESN better'' indicate significantly lower MASE for the ESN at the 5\% level.}
\label{tab:wilcoxon-quarterly}
\end{table}

\newpage

\subsection*{A.5 Computational details}
\phantomsection
\label{sec:appendix-computational}

The empirical analysis is conducted in the open-source programming
language R \citep{RCore}. The complete computational workflow is
organized in the R project
\href{https://github.com/ahaeusser/forecast_engine}
{\lstinline|forecast_engine|}, which contains the scripts for data
preparation, construction of the \textit{Parameter} and
\textit{Forecast} datasets, the ESN hyperparameter sweep, the random
initialization analysis, benchmark estimation, forecast evaluation, and
the generation of tables and figures.

The ESN models are estimated and forecast using the R package
\lstinline|echos| \citep{echos}. The package is available on
\href{https://cran.r-project.org/package=echos}{CRAN}, and its
development version is maintained in the
\href{https://github.com/ahaeusser/echos}{GitHub repository}. Supporting
functionality for time series processing, model estimation, and forecast
evaluation is provided by the R package \lstinline|tscv| \citep{tscv}.

The simple and statistical benchmark methods are primarily estimated
using the R package
\href{https://cran.r-project.org/web/packages/fable/index.html}
{\lstinline|fable|} \citep{fable}. Since TBATS is not implemented in
\lstinline|fable|, this benchmark is estimated separately using the R
package
\href{https://cran.r-project.org/web/packages/forecast/index.html}
{\lstinline|forecast|} \citep{forecast}. For the remaining statistical
methods, the underlying forecasting algorithms are largely comparable
across both packages. For example, \lstinline|fable::ARIMA()| and
\lstinline|forecast::auto.arima()| both implement the
Hyndman--Khandakar procedure for automatic ARIMA modeling
\citep{Hyndman2008b}. The MLP, RNN, and ES-RNN forecasts are taken from
the corresponding M4 benchmark and competition forecasts
\citep{Makridakis2020, Smyl2020}.

The empirical analysis was conducted using R version 4.5.2 on Windows 11
x64 (build 26200). All locally estimated models and runtime measurements
were executed on a system equipped with an Intel Core i7-10510U CPU at
1.80 GHz and 16 GB RAM. The reported runtimes in Tables
\ref{tab:metrics-monthly} and \ref{tab:metrics-quarterly} refer to model
estimation and forecast generation on the \textit{Forecast} dataset.
For the ESN, the runtime includes training, the internal random search
used to select the ridge penalty \(\lambda\), and recursive forecasting
with the selected frequency-specific ESN specification. The
hyperparameter sweep involved 4,752,000 ESN fits on the
\textit{Parameter} dataset. However, a total runtime for this
tuning phase was not recorded. The runtimes reported in Tables
\ref{tab:metrics-monthly} and \ref{tab:metrics-quarterly} therefore
refer only to fitting and forecasting with the already selected ESN
configuration on the \textit{Forecast} dataset. They should be
interpreted as deployment runtimes rather than as the total
computational cost of ESN tuning and forecasting. Computational
runtimes are not reported for MLP, RNN, and ES-RNN because their
forecasts were taken from the published M4 results. The reported
runtimes should be interpreted as indicative values because execution
times may vary between runs due to fluctuations in system load and
background processes.

\newpage
\singlespacing 
\bibliography{references}
\end{document}